\documentclass[review]{elsarticle}

\usepackage{lineno,hyperref}
\modulolinenumbers[5]

\usepackage[ruled,vlined]{algorithm2e}
\SetKwRepeat{Do}{do}{while}%

\usepackage{amsmath}
\usepackage{amssymb}
\usepackage{dsfont}
\usepackage{amsfonts}
\usepackage[T1]{fontenc}
\usepackage{yfonts}
\usepackage{slashbox}
\usepackage{xcolor}

\makeatletter
\def\mathcolor#1#{\@mathcolor{#1}}
\def\@mathcolor#1#2#3{%
  \protect\leavevmode
  \begingroup
    \color#1{#2}#3%
  \endgroup
}
\makeatother

\usepackage{graphicx}
\graphicspath{ {./figuers/} }
\usepackage{epstopdf}
\usepackage{epsfig}
\usepackage{caption}
\usepackage{subcaption}

\usepackage{longtable}
\usepackage{multicol}
\allowdisplaybreaks[4]
\usepackage{tikz}
\usepackage{tikz}
\usetikzlibrary{positioning,shadows,arrows,snakes,shapes}

\allowdisplaybreaks

\makeatletter

\newcommand{\Rmnum}[1]{\expandafter\@slowromancap\romannumeral #1@}
\makeatother

\newdefinition{rmk}{Definition}
\newproof{prf}{Proof}

\journal{}

\begin{document}

\begin{frontmatter}
%

\title{Cooperative Hierarchical Dirichlet Processes: Superposition vs. Maximization}

\author{Junyu Xuan\corref{}}
\ead{Junyu.Xuan@uts.edu.au}

\author{Jie Lu\corref{}}
\ead{Jie.Lu@uts.edu.au}

\author{Guangquan Zhang\corref{mycorrespondingauthor}}
\ead{Guangquan.Zhang@uts.edu.au}
\cortext[mycorrespondingauthor]{Corresponding author}

\author{Richard Yi Da Xu\corref{}}
\ead{Yida.Xu@uts.edu.au}

\address{Centre for Artificial Intelligence, \\
Faculty of Engineering and Information Technology, \\
University of Technology Sydney, \\
PO Box 123, Broadway, NSW 2007, Sydney, Australia}

\begin{abstract}
The cooperative hierarchical structure is a common and significant data structure observed in, or adopted by, many research areas, such as: text mining (author-paper-word) and multi-label classification (label-instance-feature). Renowned Bayesian approaches for cooperative hierarchical structure modeling are mostly based on topic models. However, these approaches suffer from a serious issue in that the number of hidden topics/factors needs to be fixed in advance and an inappropriate number may lead to overfitting or underfitting. One elegant way to resolve this issue is Bayesian nonparametric learning, but existing work in this area still cannot be applied to cooperative hierarchical structure modeling.

In this paper, we propose a cooperative hierarchical Dirichlet process (CHDP) to fill this gap. Each node in a cooperative hierarchical structure is assigned a Dirichlet process to model its weights on the infinite hidden factors/topics. Together with measure inheritance from hierarchical Dirichlet process, two kinds of measure cooperation, i.e., superposition and maximization, are defined to capture the many-to-many relationships in the cooperative hierarchical structure. Furthermore, two constructive representations for CHDP, i.e., stick-breaking and international restaurant process, are designed to facilitate the model inference. Experiments on synthetic and real-world data with cooperative hierarchical structures demonstrate the properties and the ability of CHDP for cooperative hierarchical structure modeling and its potential for practical application scenarios.
\end{abstract}


\end{frontmatter}

\linenumbers

\section{Introduction}

\begin{figure}[!t]
\begin{subfigure}{0.48\textwidth}
\centering
\begin{tikzpicture}[scale=0.6,
    fact/.style={circle , draw=none, rounded corners=1mm, fill=red, drop shadow,        text centered, anchor=north, text=red},
    state/.style={diamond, draw=none, fill=orange, circular drop shadow,
        text centered, anchor=north, text=orange},
    leaf/.style={rectangle, draw=none, fill=blue, circular drop shadow,
        text centered, anchor=north, text=blue}
]

\node (State10) at (1.5,-2) [state] {$G_{10}$} ;
\node (State11) at (-1.5,-2) [state] {$G_{11}$} ;

\node (State20) at (2,-6) [leaf] {$G_{20}$} ;
\node (State21) at (0,-6) [leaf] {$G_{21}$} ;
\node (State22) at (-2,-6) [leaf] {$G_{22}$} ;

\node (State30) at (3,-9) [fact] {$G_{30}$} ;
\node (State31) at (1,-9) [fact] {$G_{31}$} ;
\node (State32) at (-1,-9) [fact] {$G_{32}$} ;
\node (State33) at (-3,-9) [fact] {$G_{33}$} ;

\path [->,thick]	

(State10)   edge (State20)
            edge (State21)

(State11)   edge (State22)

(State20)   edge (State30)

(State21)   edge (State31)

(State22)   edge (State32)
            edge (State33)
;

\end{tikzpicture}
\caption{Type-I (Non-cooperative)}
\label{fig:nchs}
\end{subfigure}
\begin{subfigure}{0.48\textwidth}
\centering
\begin{tikzpicture}[scale=0.6,
    fact/.style={circle, draw=none, rounded corners=1mm, fill=red, drop shadow,        text centered, anchor=north, text=red},
    state/.style={diamond, draw=none, fill=orange, circular drop shadow,
        text centered, anchor=north, text=orange},
    leaf/.style={rectangle , draw=none, fill=blue, circular drop shadow,
        text centered, anchor=north, text=blue}
]

\node (State10) at (1.5,-2) [state] {$G_{10}$} ;
\node (State11) at (-1.5,-2) [state] {$G_{11}$} ;

\node (State20) at (2,-6) [leaf] {$G_{20}$} ;
\node (State21) at (0,-6) [leaf] {$G_{21}$} ;
\node (State22) at (-2,-6) [leaf] {$G_{22}$} ;

\node (State30) at (3,-9) [fact] {$G_{30}$} ;
\node (State31) at (1,-9) [fact] {$G_{31}$} ;
\node (State32) at (-1,-9) [fact] {$G_{32}$} ;
\node (State33) at (-3,-9) [fact] {$G_{33}$} ;

\path [->,thick]	

(State10)   edge (State20)
            edge (State21)
            edge (State22)

(State11)   edge (State20)
            edge (State21)
            edge (State22)

(State20)   edge (State30)
            edge (State31)
            edge (State32)
            edge (State33)

(State21)   edge (State30)
            edge (State31)
            edge (State32)
            edge (State33)

(State22)   edge (State30)
            edge (State31)
            edge (State32)
            edge (State33)
;
\end{tikzpicture}
\caption{Type-II (Cooperative)}
\label{fig:chs}
\end{subfigure}
\caption{Two types of hierarchical structures}
\label{fig:hs}
\end{figure}
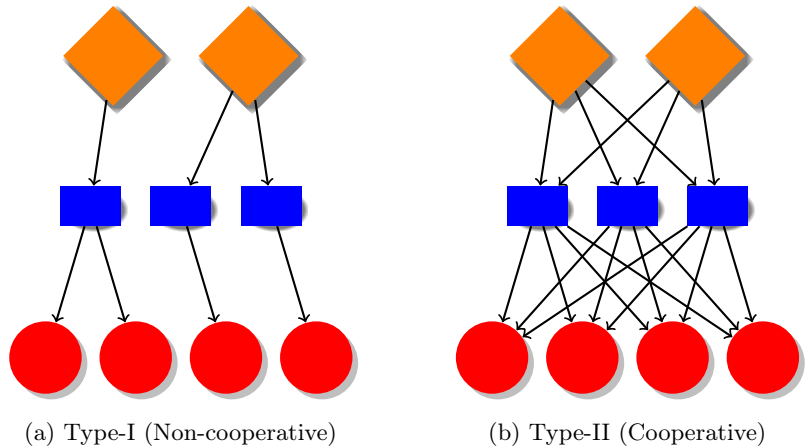

A hierarchical structure has multiple layers, and each layer contains a number of nodes that are linked to the nodes in the higher and lower layers, as illustrated in Figure \ref{fig:hs}. This kind of structure is very common and pervasive, and has been adopted in many different sub-fields in the artificial intelligence area. One example of such structure is found in text mining. Consider all the papers in a scientific journal (e.g., \emph{Artificial Intelligence}). An \emph{author-paper-word} \cite{atm3} hierarchical structure emerges, given each \emph{author} writes and publishes a number of scientific \emph{papers} in this journal, and each \emph{paper} is composed of several different \emph{words}. Learning from \emph{author-paper-word} structure is useful for collaborators' recommendations, authors disambiguation, paper clustering, statistical machine translation \cite{Xiong201654}, and so on. Another example occurs within image processing. The \emph{scene-image-feature} hierarchical structure is formed because each \emph{image} may belong to several \emph{scenes}, such as beach or urban \cite{Boutell20041757}, and an image is also described by an abundance of \emph{features}, such as grayscale and texture. Learning from \emph{scene-image-feature} structure could at least benefit image search and context-sensitive image enhancement.


Current state-of-the-art Bayesian approaches to learn from this hierarchical structure are mainly based on topic models \cite{blei2003latent,7015568} that are a kind of probabilistic graphical models \cite{Flach2001199} and were originally designed for modeling a two-level hierarchical structure: \emph{document-word}.
Their basic idea is to construct a Bayesian prior based on manipulations on probabilistic distributions, e.g., Dirichlet and Multinomial distributions \cite{Blei:2012:PTM}, to map documents and words into a latent topic space.
For example, papers in the \emph{Artificial Intelligence Journal} cover multiple research topics, such as \emph{machine learning}, \emph{intelligent robotics}, \emph{case-based reasoning}, and \emph{knowledge representation}. Each paper in this journal could be seen as a combination of these research topics, and each topic is described by a weighted word vector. Beyond the two-level hierarchical structure, some three-level hierarchical structures have also been successfully modelled by incorporating additional document side information, such as: \emph{author-document-word} \cite{atm3}, \emph{emotion-document-word} \cite{emotion}, \emph{entry-document-word} \cite{entry} and \emph{label-document-word} \cite{labels}.


A major issue in existing (parametric) topic model-based hierarchical structure modeling is that the hidden topic number in the defined priors needs to be fixed in advance. This number is usually chosen with domain knowledge. After fixing the number of topics, Dirichlet, multinomial, and other fixed-dimensional distributions could be adopted as the building blocks for (parametric) topic models. However, discovering an appropriate number is very difficult and sometimes unrealistic in many real-world applications. For example, limiting any given corpus to a fixed exact number of topics is apparently unrealistic. Furthermore, this may lead to overfitting where there are too many topics, so that relatively specific topics will not generalise well to unseen observations; Underfitting is the opposite case, where there are too few topics, so unrelated observations will be assigned to the same topic \citep{6784083}. This number is supposed to be inferred from the data, i.e., let the data speak. A number of methods can be used to nominate the number of topics, including cross-validation techniques \citep{griffiths2004finding}, but they are not efficient because the algorithm has to be restarted a number of times before determining the optimal number of topics \citep{griffiths2004finding,6784083}.


One elegant approach to resolve the above issue is \emph{Bayesian nonparametric learning} - a key approach for learning the number of mixtures in a mixture model (also called the model selection problem) \cite{Gershman20121}. The idea of Bayesian nonparametric learning is to use stochastic processes to replace the traditional fixed-dimensional probability distributions. The merit of these stochastic processes is that they have a theoretically infinite number of factors\footnote{We do not distinguish \emph{factor} with \emph{topic} throughout this paper.} and let the data determine the used number of factors. Many probabilistic models with fixed dimensions have been extended to infinite ones with the help of stochastic processes. One typical example is the famous Gaussian mixture model, which was extended into an infinite Gaussian mixture model \cite{rasmussen1999infinite} using the Dirichlet process. As for hierarchical structure modeling, the hierarchical Dirichlet process (HDP) \cite{teh2006hierarchical} is the most well known, which uses the relationship between a stochastic process and its base measure to capture the hierarchical structure in data: more details are given in the preliminary knowledge section. Due to its success, many extensions have been developed to account for different situations, such as: a supervised version \cite{6784083} for modeling additional labels and an incremental version \cite{6137314} for streaming data.


However, this state-of-the-art HDP-based work can only model one special type of hierarchical structure, however there are actually two types, as shown in Figure \ref{fig:hs}, which are distinguished by the number of parent nodes for each node. In Type-I hierarchical structures, as illustrated in Figure \ref{fig:nchs}, each node has one and only one parent node which could be seen as a group, and in turn is assigned to higher level groups. In Type-II hierarchical structures, as illustrated in Figure \ref{fig:chs}, each node may have more than one parent node. In this paper, we term this structure a \emph{cooperative hierarchical structure}. Type-II is typically considered more general than Type-I, because Type-I can be seen as a special case of Type-II. Note that the renowned hierarchical Dirichlet process and its extensions (e.g., HDP-HMM \cite{Wulsin201455}, HDP-based hierarchical distance-dependent Chinese Restaurant process (hddCRP) \cite{DBLP:conf/uai/GhoshRSS14}, and HDP-based scene detection \cite{DBLP:conf/ijcai/MitraBB15}) are all particularly designed after Type-I hierarchical structures but fail to model Type-II hierarchical structures. Consider the former example on an \emph{author-paper-word} structure. Using a Type-I hierarchical structure for the text mining area would, in this case, imply that each \emph{paper} was only written by one \emph{author}. This applies to \emph{scene-image-feature} structures as well. Despite a certain rationality in some situations, the constraints of the Type-I hierarchical structure are too restrictive to model many real-world phenomena, so a new Bayesian nonparametric prior is a must for modeling Type-II hierarchical structures.


This paper proposes a Bayesian nonparametric model for cooperative hierarchical structures, based on the renowned hierarchical Dirichlet process (HDP), which we call the \emph{cooperative hierarchical Dirichlet process}(CHDP). More specifically, it is built on two operations for random measures from the Dirichlet process: \emph{Inheritance} from the hierarchical Dirichlet process; \emph{Cooperation}, an innovation proposed in this paper, to account for multiple parent nodes in Type-II hierarchical structures. More specially, we have designed two mechanisms for \emph{Cooperation}: one is \emph{Superposition} and the other is \emph{Maximization}. Based on these operations, we propose the \emph{cooperative hierarchical Dirichlet process} along with its two constructive representations. Although the proposed CHDP elegantly captures cooperative hierarchical structures, it also brings additional challenges to model inference. To resolve this challenge, we introduce two inference algorithms based on the proposed two representations. Experiments on synthetic and  real-world tasks show the properties of the proposed CHDP and its usefulness in cooperative hierarchical structure modeling.


In summary, the main two contributions of this article are as follows:
\begin{itemize}
  \item we innovatively propose a cooperative hierarchical Dirichlet process based on operations on random measures: \emph{Inheritance}, \emph{Cooperation: Superposition} and \emph{Cooperation: Maximization}, which can be used to model the cooperative hierarchical structures that cannot be modelled by existing Bayesian nonparametric models;
  \item two constructive representations (i.e., the international restaurant process and stick-breaking) and the corresponding inference algorithms for the cooperative hierarchical Dirichlet process are proposed to facilitate model inference, which rise to the challenge brought about by \emph{Inheritance}, \emph{Cooperation: Superposition} and \emph{Cooperation: Maximization} between the random measures.
\end{itemize}


The remainder of this article is organized as follows. Section 2 discuses related work. The definitions and constructive representations of the DP and the HDP, which are the preliminary knowledge of the proposed model, are reviewed in Section 3. The CHDP and its two constructive representations are presented in Section 4 with two corresponding inference algorithms in Section 5. Section 6 evaluates the properties of CHDP and conducts comparative experiments on real-world tasks. Section 7 concludes this study and discusses possible future work.

\section{Related work}

This section reviews the study on hierarchical structures using Bayesian nonparametric models. We organize the existing work in this area into two groups: one group aims to learn \emph{out} a hierarchical structure from (plain) data; the other group aims to learn \emph{from} data with a hierarchical structure. Although the two groups are similar, they are developed for different situations: the input of the first group is a plain dataset (e.g., a collection of documents or images) and the output is a hieratical structure; the input of the second group is a hierarchical data structure and the output is a new hidden factor space. Our study in this paper is within the second group.

\subsection{Learning \emph{out} hierarchical structures using Bayesian nonparametrics}

Hierarchical structures play an important role in machine learning because they are pervasively applied and reflect the human habit to organize information, so learning out a hierarchical structure from plain data attracts a lot of attention from researchers in the Bayesian nonparametric field. Compared to other efforts on this task, Bayesian nonparametric models have the advantage that the learned hierarchical structure is more flexible which means there is no bound of depth and/or width, making it easy to incorporate the newly arrived data.

\emph{nCRP-based.} A tree is viewed as a nested sequence of partitions by the nested Chinese restaurant process (nCRP) \cite{griffiths2004hierarchical,ncrp2010}, where a measurable space is first partitioned by a CRP \cite{blackwell1973ferguson} and each area in this partition is further partitioned into several areas using CRP. In this way, a tree with infinite depth and branching can be generated. A datum (e.g., a document) is associated with a path in the tree using DP by nCRP \cite{ncrp2010} or a flexible Martingale \cite{steinhardt2012flexible} prior, and it can associate with a subtree of the generated tree using the HDP \cite{teh2006hierarchical} prior in the nested HDP \cite{nhdp2015} instead of a path.

\emph{Stick-breaking-based.} It is known that the traditional stick-breaking process \cite{sethuraman1994constructive} can infer an infinite set, and it has also been revised to infer an infinite tree structure. An iterative stick-breaking process is used to construct a Polya tree (PT) \cite{10.2307/2242009} in a nested fashion, and a datum is associated with a leaf node of the generated tree. The traditional stick-breaking process is revised to generate breaks with a tree structure and results in tree structured stick-breaking (TSSB) \cite{ghahramani2010tree} where a datum is attached to a node in the generated tree.

\emph{Diffusion-based.} This kind of method holds the idea that data are generated by a diffusion procedure with several divergences during this procedure and additional time varying continuous stochastic processes (i.e., Markov process) are needed for divergence control. A datum is placed at the end of the branches of diffusions. Both Kingman's coalescent \cite{teh2009bayesian,KINGMAN1982235,teh2011modelling} and the Dirichlet diffusion tree (DDT) \cite{neal2003density} define a prior for an infinite (binary) tree. DDT is extended to a more general structure: multifurcating branches by the Pitman-Yor diffusion tree (PYDT) \cite{knowles2011pitman,6777276} and to feature hierarchy by the beta diffusion tree (BDT) \cite{heaukulani2014beta}.

\emph{Other.} Motivated by the deep belief network (DBN) \cite{hinton2006fast}, the Poisson gamma belief network (PGBN) \cite{zhou2015gamma} is proposed to learn a hierarchical structure where nodes have nonnegative real-valued weights rather than binary-valued weights in DBN and the width of each layer is flexible rather than fixed. Each layer node can be seen as an abstract feature expression of the input data.

To summarize, a variety of excellent work has been proposed in this direction, but this is beyond the scope of this work.

\subsection{Learning \emph{from} hierarchical structures using Bayesian nonparametrics}

The most well-known and significant Bayesian nonparametric model for learning from hierarchical structures is the hierarchical Dirichlet process (HDP) \cite{teh2006hierarchical}, which is based on layering DPs. Each node in the hierarchical structure is assigned a DP, and the relationship between nodes is modeled by the relation between a DP and its base measure. Due to its success, many extensions have been developed to account for different situations: supervised HDP \cite{6784083} is proposed to incorporate additional label information of hierarchical structures; dynamic HDP \cite{dynamichdp,Zhang:2010} is used to model the time-varying change of hierarchical structures; incremental HDP \cite{6137314} is for streaming hierarchical structures; the tree extension of HDP \cite{Canini:2011} and the combination with deep Boltzmann Machine (DBM) \cite{salakhutdinov2009deep} are used to learn out a different level of abstract features \cite{6389680}; and the adapted HDP \cite{6916915} can fuse multiple heterogeneous aspects.

A similar idea was adopted in the gamma-negative binomial process \cite{nbpcm,7373353}, beta-negative binomial process \cite{6802382}, hierarchical beta process \cite{hbp} and hierarchical Poisson models \cite{hpm}. Different stochastic processes, e.g., beta, Gamma, Poisson and negative binomial processes, used in these models are piled to account for different kinds of data (i.e., binary or count data) in the hierarchical structure. Note that these models can also be used to learn out a hierarchical structure if the hidden layers are fixed in advance for plain data.

To summarize, current state-of-the-art research in this group is mostly based on the hierarchical idea originally designed in HDP, so they can only be applied to Type-I hierarchical structures, as discussed in the introduction.

\section{Preliminary knowledge}

The CHDP is built on two existing Bayesian nonparametric priors: the Dirichlet process (DP) and the hierarchical Dirichlet process (HDP). In this section, we review their definitions and constructive representations that have been used to understand and build the proposed CHDP in the following section. Some important notations used throughout this paper are summarized in Table \ref{notations}.

\begin{footnotesize}
\begin{longtable}[t]{c|p{10cm}}
\caption{Important notations in this paper} \label{notations}\\
\hline
Symbols & Description
\\\hline
$\Theta$                & a measurable space
\\\hline
$G$                     & a random measure from DP
\\\hline
$G_0$/$G^1_0$           & global random measure from DP at the first layer
\\\hline
$G_a$/$G^2$             & a random measure from DP at the second layer
\\\hline
$G_d$/$G^3$             & a random measure from DP at the third layer
\\\hline
$G_i^\ell$              & $i$-th random measure from DP at the $\ell$-th layer
\\\hline
$N^{\ell}$              & the number of random measures at $\ell$-th layer
\\\hline
$H$                     & base measure of DP
\\\hline
$\gamma$                & the parameter of $H$ (when it is a Dirichlet distribution)
\\\hline
$\Omega$                & a random partition
\\\hline
$\Omega_k$              & a measurable set in a random partition
\\\hline
$k$                     & an index of a measurable set/partition/factor/topic/dish
\\\hline
$K$                     & the number of measurable sets in a partition/factors/topics/dishes
\\\hline
$a$                     & a chef/node at the second layer
\\\hline
$A$                     & number of chefs/nodes at the second layer
\\\hline
$d$                     & a restaurant
\\\hline
$D$                     & number of restaurants/nodes at the third layer
\\\hline
$t$                     & a table in a restaurant
\\\hline
$T_{d}$                 & the table number in restaurant $d$
\\\hline
$T^a_{d}$               & the table number in restaurant $d$ served by chef $a$
\\\hline
$T_{a,o}$               & the number of tables served by menu option $o$ of chef $a$
\\\hline
$T_{k}$                 & the number of tables served by dish $k$
\\\hline
$o$                     & a menu option on the personal menu
\\\hline
$O_a$                   & the number of menu options on the personal menu of chef $a$
\\\hline
$O_k$                   & the number of menu options with dish name $k$
\\\hline
$V$                     & the number of different words in a corpus
\\\hline
$\theta_k$              & $k$-th partition/factor/topic/dish of DP (one point in $\Theta$)
\\\hline
$\theta_{a,o}$          & assigned factor to menu option $o$ of chef $a$
\\\hline
$\theta_{d,t}$          & assigned factor to table $t$ in restaurant $d$
\\\hline
$\theta_{d,n}$/$\theta_{d,i}$          & assigned factor to data/customer $n$/$i$ in restaurant $d$
\\\hline
$\alpha$                & concentration parameter of general DP
\\\hline
$\alpha_0$              & concentration parameter of global DP at first layer
\\\hline
$\alpha_a$              & concentration parameter of DPs at second layer
\\\hline
$\alpha_d$              & concentration parameter of DPs at third layer
\\\hline
$\nu_{k}$               & $k$-th stick break from beta distribution $Beta(1, \alpha)$
\\\hline
$\pi_k$                 & the stick weight of $k$-th atom/factor from general DP
\\\hline
$\nu_{0,k}$             & $k$-th stick break from beta distribution $Beta(1, \alpha_0)$
\\\hline
$\pi_{0,k}$             & the stick weight of $k$-th atom/factor from global DP at first layer
\\\hline
$\nu_{a,o}$             & $o$-th stick break from beta distribution $Beta(1, \alpha_a)$
\\\hline
$\pi_{a,o}$             & the stick weight of $o$-th atom/factor from DP at second layer
\\\hline
$\nu_{d,t}$             & $t$-th stick break from beta distribution $Beta(1, \alpha_d)$
\\\hline
$\pi_{d,t}$             & the stick weight of $t$-th atom/factor from DP at third layer
\\\hline
$z_{a,o}$               & the assigned index of factor/dish of a node at first layer for a option $o$ of $a$
\\\hline
$z_{d,t}$               & the assigned index of factor/option of a node at second layer for a table $t$ of $d$
\\\hline
$z_{d,n}$               & the assigned index of factor/table of a node at third layer for a data $n$ of $d$
\\\hline
$N_{d}$                 & the number of data/customers in restaurant $d$
\\\hline
$N_{d,t}$               & the number of data/customers sitting at table $t$ of  restaurant $d$
\\\hline
$N^a_{d,t}$             & the number of data/customers sitting at table $t$ of  restaurant $d$ served by chef $a$
\\\hline
$u_{0,k}, r_{0,k}$             & the variational parameters for stick breaks at the top layer
\\\hline
$u_{a,o}, r_{a,o}$             & the variational parameters for stick breaks at the second layer
\\\hline
$u_{d,t}, r_{d,t}$             & the variational parameters for stick breaks at the third layer
\\\hline
$\varsigma_{a,o}$             & the variational parameters for $z_{a,o}$
\\\hline
$\varsigma_{d,t}$             & the variational parameters for $z_{d,t}$
\\\hline
$\varsigma_{d,n}$             & the variational parameters for $z_{d,n}$
\\\hline
$\vartheta_{k}$             & the variational parameter for $\theta_k$
\\\hline
\end{longtable}
\end{footnotesize}

\subsection{Dirichlet process}

The Dirichlet process \cite{ferguson1973bayesian,DP2010} is the pioneer and foundation of Bayesian nonparametric learning. Its definition is as follows:
\begin{rmk} [Dirichlet Process]\label{def:dp}
A \emph{Dirichlet process (DP)} \cite{ferguson1973bayesian,DP2010}, which is specified by a base measure $H$ on a measurable space $\Theta$ and a concentration parameter $\alpha$, is a set of countably infinite random variables that can be seen as the measures on measurable sets from a random infinite partition $\{\Omega_k\}_{k=1}^{\infty}$ of $\Theta$. For any finite partition $\{\Omega_k\}_{k=1}^{K}$, the variables (measures on these measurable sets) from DP satisfy a Dirichlet distribution parameterized by the measures from the base measure $H$
\begin{equation*}
\begin{aligned}
(G(\Omega_1), G(\Omega_2), \ldots, G(\Omega_K) ) \sim Dir(\alpha H(\Omega_1), \alpha H(\Omega_2), \ldots, \alpha H(\Omega_K))
\end{aligned}
\end{equation*}
where $G$ is a realization of $DP(\alpha, H)$ and $Dir()$ denotes the Dirichlet distribution.
\end{rmk}

Since $G$ is a discrete measure with probability one \cite{ferguson1973bayesian}, the mass $G(\Omega_k)$ will concentrate on one point (i.e., $\theta_k \in \Omega_k$, called a topic/a factor/an atom\footnote{We do not distinguish these terms throughout this paper.} in this paper) of $\Omega_k$, so an alternative definition of $G$ is
\begin{equation}\label{dpdefintion2}
\begin{aligned}
G = \sum_{k=1}^{\infty} \pi_k \delta_{\theta_k},~~\sum_{k=1}^{\infty} \pi_k = 1,~~\theta_k \sim H
\end{aligned}
\end{equation}
where $\{\theta_k\}_{k=1}^{\infty}$ denotes countable infinite points in measurable space $\Theta$ and are sampled according to the base measure $H$; $\pi_k = G(\Omega_k)$ is the measure value from $G$ on a measurable set $\Omega_k$ and it can be seen as the (normalized) weight of $\theta_k$ in $\{\theta_k\}_{k=1}^{\infty}$; $\delta_{\theta_k}$ is a Dirac measure parameterized by $\theta_k$ (i.e., $\delta_{\theta_k}(\hat{\theta}) = 1$ if $\hat{\theta} = \theta_k$; 0, otherwise). One draw from $G$ would be one of $\{\theta_k\}_{k=1}^{\infty}$ according to their relative weights $\{\pi_k\}_{k=1}^{\infty}$.

Considering its infinite and discrete nature, DP is commonly adopted as the prior for mixture models \cite{rasmussen1999infinite}, such as:
\begin{equation}\label{dpdefintion3}
\begin{aligned}
x_i \sim F(\theta_i), ~~ \theta_i \sim G
\end{aligned}
\end{equation}
where $x_i$ is a data point generated according to a distribution $F()$ parameterized by a draw $\theta_i$ from $G$. Due to the discrete nature of $G$, we have $\theta_i \in \{\theta_k\}_{k=1}^{\infty}$ with the implication of \emph{data clustering} according to their assigned $\theta_i$. For  computational convenience, $F()$ is normally set as a multinomial distribution because it is conjugate with Dirichlet distribution. Document modeling is a successful application of this mixture model: $\theta_k$ is a $V$-dimensional (normalized) vector (named a topic) where $V$ is the number of different words in a text corpus.

In Bayesian posterior analysis of DP, a representation of $G$ from a DP is needed. According to whether $G$ is represented explicitly or not, there are two kinds of constructive representations: Chinese restaurant process (CRP) representation and stick-breaking representation.

\subsubsection{Chinese restaurant process (CRP) representation}
\label{sec:dp:crp}

A marginal constructive representation is the Chinese restaurant process \cite{blackwell1973ferguson}, which directly generates $\theta_i$ for the $i$-th data point (they are exchangeable) with $G$ marginalized out as follows:
\begin{equation}\label{dpcrp}
\begin{aligned}
\theta_i | \theta_1,\cdots, \theta_{i-1} \sim \sum_{j=1}^{i-1} \frac{1}{\alpha+i-1}\delta_{\theta_j} + \frac{\alpha}{\alpha+i-1}H
\end{aligned}
\end{equation}
where $\frac{1}{\alpha+i-1}$ is the probability of taking the previous ones and $\frac{\alpha}{\alpha+i-1}$ is the probability of taking a new one according to $H$. Here, the weights $\pi_k$ in Eq. (\ref{dpdefintion2}) are implicitly reflected by the ratio of $\theta_k$ in $\{\theta_i\}_{i \rightarrow \infty}$.

The name comes from a metaphor used to understand Eq. (\ref{dpcrp}). In a Chinese restaurant, the $i$-th customer walks into this restaurant and chooses to sit at an occupied table with the probability $\frac{1}{\alpha+i-1}$ or a new table with the probability $\frac{\alpha}{\alpha+i-1}$. If the customer picks an occupied table, she eats the dish already on the table; if a new table is picked, she needs to order a new dish for the table from $H$. As a result, $\theta_i$ is the dish eaten by the $i$-th customer.

\subsubsection{Stick-breaking representation}

Another explicit way (named stick-breaking) to construct $G$ is proposed in \cite{sethuraman1994constructive} as follows
\begin{equation*}
\begin{aligned}
G = \sum_k^{\infty} \nu_k \prod_{j=1}^{k-1} (1-\nu_j) \delta_{\theta_k},~\nu_k \sim Beta(1, \alpha),~ \theta_k \sim H
\end{aligned}
\end{equation*}
where $Beta()$ denotes a Beta distribution and $\nu_k$ is the $k$-th random break from a unit stick with Beta distribution parameterized by $1$ and $\alpha$. We can see that the weights $\pi_k$ in Eq. (\ref{dpdefintion2}) can be explicitly represented by $\nu_k \prod_{j=1}^{k-1} (1-\nu_j)$.

\subsection{Hierarchical Dirichlet processes}

The hierarchical Dirichlet process \cite{teh2006hierarchical} is built by piling a DP above another DP through an elegant method that can share the factors across the hierarchical structure. Its definition is as follows:
\begin{rmk} [Hierarchical Dirichlet Process]\label{def:hdp}
A \emph{hierarchical Dirichlet process (HDP)} \cite{teh2006hierarchical} is a distribution over a set of random probability measures over $\Theta$. The process defines a set of random probability measures $\{G_d\}^D_{d=1}$ and a global random probability measure $G_0$. The global measure $G_0$ is distributed as a Dirichlet process parameterized by a concentration parameter $\alpha$ and a base (probability) measure $H$
\begin{equation*}
\begin{aligned}
G_0 \sim DP(\alpha, H)
\end{aligned}
\end{equation*}
Each random measure $G_d$ is conditionally independent from the others given $G_0$, and is also distributed as a Dirichlet process with the parameter $\alpha_d$ and a base probability measure $G_0$
\begin{equation*}
\begin{aligned}
G_d \sim DP(\alpha_d, G_0)
\end{aligned}
\end{equation*}
\end{rmk}

This definition actually defines an operation between two DPs which will be discussed in more detail in the following section. It was originally designed to model \emph{group data}. For example, there are $D$ documents (i.e., groups) and each $G_d$ could be adopted to model one document using the mixture idea in Eq. (\ref{dpdefintion3}). Note that extending the above two-layer HDP to more layers is straightforward under this definition.

Analogous to DP, the representation for HDP is also required for model inference. There are two candidates: Chinese restaurant franchise representation and stick-breaking representation.

\subsubsection{Chinese restaurant franchise (CRF) representation}
\label{sec:hdp:crf}

Similar to the CRP for DP, HDP has its own marginal representation with $G_0$ and $\{G_d\}_{d=1}^D$ marginalized out (named the Chinese Restaurant Franchise) as follows:
\begin{equation}\label{hdpcrf}
\begin{aligned}
\theta_{d,t} | \theta_{1,1},\cdots, \theta_{D,t-1} &\sim \sum_{k=1}^{K} \frac{T_k}{\alpha+\sum_k T_k}\delta_{\theta_k} + \frac{\alpha}{\alpha+\sum_k T_k}H
\\
\theta_{d,i} | \theta_{d,1},\cdots, \theta_{d,i-1} &\sim \sum_{t=1}^{T_d} \frac{N_{d,t}}{\alpha_d+i-1}\delta_{\theta_{d,t}} + \frac{\alpha_d}{\alpha_d+i-1}G_0
\end{aligned}
\end{equation}
where $T_k$ denotes the number of $\theta_{d,t}$ associated with $\theta_k$ and $N_{d,t}$ denotes the number of $\theta_{d,i}$ associated with $\theta_{d,t}$ in $d$. Note that although $G_0$ appears in the above representation, we do not need to represent it explicitly as we can use the first line of Eq. (\ref{hdpcrf}) when we need to sample from $G_0$ in second line of Eq. (\ref{hdpcrf}).

The metaphor for CRF in Eq. (\ref{hdpcrf}) is as follows \cite{teh2006hierarchical}. There are $D$ Chinese restaurants with a shared menu. The $i$-th customer walks into the $d$-th restaurant and picks an occupied table at which to sit with the probability $\frac{N_{d,t}}{\alpha_d+i-1}$ or a new table with the probability $\frac{\alpha_d}{\alpha_d+i-1}$. If this customer picks an occupied table, she just eats the dish already on that table; if a new table is picked, she needs to order a new dish. The new dish is ordered from the menu according to its popularity. The probability that the new dish is the same as the one on other tables has a probability of $\frac{T_k}{\alpha+\sum_k T_k}$ and the probability that it is a new dish is $\frac{\alpha}{\alpha+\sum_k T_k}$, where $T_k$ is the number of tables with the same dish $\theta_k$. As a result, $\theta_{d,t}$ is the dish on table $t$ of restaurant $d$, and $\theta_{d,i}$ is the dish eaten by customer $i$ in restaurant $d$.

\subsubsection{Stick-breaking representation}

As for stick-breaking-based representation, there are two versions \cite{teh2006hierarchical,sethuraman1994constructive} for HDP. In this paper, we adopt the Sethuraman's version \cite{sethuraman1994constructive,wang2011online} (with two layer) as follows:
\begin{equation*}
\begin{aligned}
G_0 &= \sum_k^{\infty} \pi_{0,k} \delta_{\theta_k} & \pi_{0,k}&=\nu_{0,k} \prod_{j=1}^{k-1} (1-\nu_{0,j}) & \nu_{0,k} &\sim Beta(1, \alpha_0)
\\
G_d &= \sum_t^{\infty} \pi_{d,t} \delta_{\theta_{d,t}} & \pi_{d,t}&=\nu_{d,t} \prod_{j=1}^{t-1} (1-\nu_{d,j}) & \nu_{d,t} &\sim Beta(1, \alpha_d)
\\
\theta_k &\sim H & \theta_{d,t} &= \theta_{z_{d,t}} & z_{d,t} &\sim \pi_0
\end{aligned}
\end{equation*}
where $z_{d,t}$ denotes an index to one of $\{\theta_k\}_{k=1}^{\infty}$. Sethuraman's version has an advantage in that the stick weights at different layers are decoupled which makes the posterior inference easier. From this constructive representation, we can see the factor sharing property of HDP. The $G_d$ at the lower layer shares the factors $\{\theta_k\}_{k=1}^{\infty}$ of $G_0$ at higher layers. Another interesting point is that the constructions of $\pi_{0}$ and $\{\pi_d\}$ are independent and the only connections between $G_0$ and $\{G_d\}$ are the relationships between $\theta_{k}$ and $\{\theta_{d}\}$.

\section{Cooperative hierarchical Dirichlet processes}

As discussed in the Introduction, there are two types of hierarchical structures. In this section, we formally define and model the second type: the cooperative hierarchical structure.

\begin{rmk}[Cooperative Hierarchical Structure]\label{def:chs}
A \emph{cooperative hierarchical structure (CHS)}, as illustrated in Figure \ref{fig:chs}, is composed of nodes assigned to different layers. Each node in the structure may link to multiple parent nodes and child nodes.
\end{rmk}

A real-world example of CHS is: \emph{author-paper-word} data. This data has three-layer nodes: nodes in first layer denote \emph{authors}; nodes in the second layer denote \emph{papers}; nodes in the third layer denote \emph{words}. If an \emph{author} writes a \emph{paper}, there is a link between two corresponding nodes; similarly, there is a link between a \emph{paper} and a \emph{word} if this paper contains this word.

Note that there is an implicit assumption of HDP in Definition 2 that each node can only have one parent node, so HDP fails to model CHS.
To capture CHS, we first formally define three operations on random measures from DP as follows:
\begin{rmk} [Inheritance]
A probability measure $G_1$ is the \emph{Inheritance} from another probability measure $G_2$ from DP on space $\Theta$ by taking $G_2$ as its base measure
\begin{equation*}
\begin{aligned}
G_1 \sim DP(\alpha_1, G_2),~~~G_2 \sim DP(\alpha_2, H)
\end{aligned}
\end{equation*}
where $\alpha_1$ and $\alpha_2$ are DP parameters. The discrete nature of $G_2$ enables $G_1$ to inherit factors/atoms from $G_2$.
\end{rmk}

Note that this operation is a more formal definition than the one in Definition \ref{def:hdp}.

\begin{rmk} [Cooperation: Superposition]\label{def:superposition}
A measure $G$ is the \emph{Superposition} of two probability measures, i.e., $G_1$ and $G_2$, from DP on the same space $\Theta$,  if
\begin{equation*}
\begin{aligned}
G = G_1 \oplus G_2
\end{aligned}
\end{equation*}
where $G$ is a new probability measure on space $\Theta$ and $\oplus$ denotes the convex combination. For any given partition $\{\Omega\}_{k=1}^{\infty}$ on $\Theta$, it has
\begin{equation*}
\begin{aligned}
G(\Omega_k) = \frac{G_1(\Omega_k) + G_2(\Omega_k)}{\sum_k (G_1(\Omega_k) + G_2(\Omega_k))}
\end{aligned}
\end{equation*}
Extending the \emph{Superposition} of more than two probability measures is straightforward.
\end{rmk}

\begin{rmk} [Cooperation: Maximization]\label{def:maximization}
A measure $G$ is the \emph{Maximization} of two probability measures, i.e., $G_1$ and $G_2$, from DP on the same space $\Theta$,  if
\begin{equation*}
\begin{aligned}
G = G_1 \vee G_2
\end{aligned}
\end{equation*}
where $G$ is a new probability measure on the space $\Theta$ and $\vee$ that is a Zadeh operator borrowed from fuzzy logic which denotes the maximization\footnote{Here, $\vee$ is a little different from its original definition, because there will be normalization after taking the maximum.}. For any given partition $\{\Omega\}_{k=1}^{\infty}$ on $\Theta$, it has
\begin{equation*}
\begin{aligned}
G(\Omega_k) = \frac{\max \{G_1(\Omega_k), G_2(\Omega_k)\}}{\sum_k \max \{G_1(\Omega_k), G_2(\Omega_k)\}}
\end{aligned}
\end{equation*}
Extending the \emph{Maximization} of more than two probability measures is also straightforward.
\end{rmk}

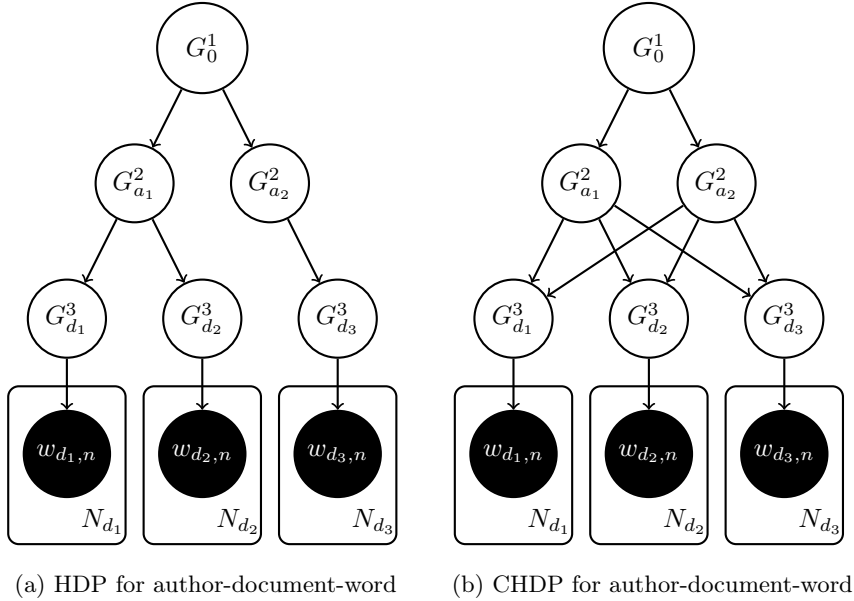
\begin{figure}[!t]
\begin{subfigure}{0.48\textwidth}
\centering
\begin{tikzpicture}[scale=0.6,
    statee/.style={circle, minimum size=1.2cm, draw=black, thick, text centered, text=black},
    fact/.style={circle, minimum size=1cm, draw=black, thick, text centered, fill=black, text=white},
    state/.style={circle, minimum size=1cm, draw=black, thick, text centered, text=black},
    leaf/.style={circle, minimum size=1cm, draw=black, thick, text centered, text=black}
]
\node (State0) at (0,0) [statee] {$G^1_{0}$} ;

\node (State11) at (1.5,-3) [state] {$G^2_{a_2}$} ;
\node (State10) at (-1.5,-3) [state] {$G^2_{a_1}$} ;

\node (State22) at (3,-6) [leaf] {$G^3_{d_3}$} ;
\node (State21) at (0,-6) [leaf] {$G^3_{d_2}$} ;
\node (State20) at (-3,-6) [leaf] {$G^3_{d_1}$} ;

\draw[thick, rounded corners] (-4.3,-7.5) rectangle (-1.7,-11);
\node (State30) at (-3,-9) [fact] {$w_{d_1,n}$} ;
\draw (-2.2, -10.5) node[scale=1] {$N_{d_1}$};

\draw[thick, rounded corners] (-1.3,-7.5) rectangle (1.3,-11);
\node (State31) at (0,-9) [fact] {$w_{d_2, n}$} ;
\draw (0.8, -10.5) node[scale=1] {$N_{d_2}$};

\draw[thick, rounded corners] (4.3,-7.5) rectangle (1.7,-11);
\node (State32) at (3,-9) [fact] {$w_{d_3, n}$} ;
\draw (3.8, -10.5) node[scale=1] {$N_{d_3}$};

\path [->, thick]	

(State0)    edge (State10)
            edge (State11)

(State10)   edge (State20)
            edge (State21)

(State11)   edge (State22)

(State20)   edge (State30)
(State21)   edge (State31)
(State22)   edge (State32)

;
\end{tikzpicture}
\caption{HDP for author-document-word}
\label{fig:hdpforchs}
\end{subfigure}
\begin{subfigure}{0.48\textwidth}
\centering
\begin{tikzpicture}[scale=0.6,
    statee/.style={circle, minimum size=1.2cm, draw=black, thick, text centered, text=black},
    fact/.style={circle, minimum size=1cm, draw=black, thick, text centered, fill=black, text=white},
    state/.style={circle, minimum size=1cm, draw=black, thick, text centered, text=black},
    leaf/.style={circle, minimum size=1cm, draw=black, thick, text centered, text=black}
]
\node (State0) at (0,0) [statee] {$G^1_{0}$} ;

\node (State11) at (1.5,-3) [state] {$G^2_{a_2}$} ;
\node (State10) at (-1.5,-3) [state] {$G^2_{a_1}$} ;

\node (State22) at (3,-6) [leaf] {$G^3_{d_3}$} ;
\node (State21) at (0,-6) [leaf] {$G^3_{d_2}$} ;
\node (State20) at (-3,-6) [leaf] {$G^3_{d_1}$} ;

\draw[thick, rounded corners] (-4.3,-7.5) rectangle (-1.7,-11);
\node (State30) at (-3,-9) [fact] {$w_{d_1,n}$} ;
\draw (-2.2, -10.5) node[scale=1] {$N_{d_1}$};

\draw[thick, rounded corners] (-1.3,-7.5) rectangle (1.3,-11);
\node (State31) at (0,-9) [fact] {$w_{d_2, n}$} ;
\draw (0.8, -10.5) node[scale=1] {$N_{d_2}$};

\draw[thick, rounded corners] (4.3,-7.5) rectangle (1.7,-11);
\node (State32) at (3,-9) [fact] {$w_{d_3, n}$} ;
\draw (3.8, -10.5) node[scale=1] {$N_{d_3}$};

\path [->, thick]	

(State0)    edge (State10)
            edge (State11)

(State10)   edge (State20)
            edge (State21)
            edge (State22)

(State11)   edge (State20)
            edge (State21)
            edge (State22)

(State20)   edge (State30)
(State21)   edge (State31)
(State22)   edge (State32)

;
\end{tikzpicture}
\caption{CHDP for author-document-word}
\label{fig:chdpforchs}
\end{subfigure}
\caption{Comparison between graphical models of HDP and CHDP for a particular hierarchical structure: author-document-word, where this simple data includes three documents written by two authors and each document $d$ has with $N_d$ words. In HDP, each document can only have one author; in CHDP, each document can have multiple authors.}
\label{fig:hdpandchdp}
\end{figure}

The defined \emph{Superposition} and \emph{Maximization} are two cooperation mechanisms between random measures, and they are not interchangeable. With the help of two mechanisms, we can model the many-to-many relationship of CHS defined in Definition \ref{def:chs}. Next, we define a new Bayesian nonparametric prior to model CHS as follows:
\begin{rmk} [Cooperative Hierarchical Dirichlet Process]\label{def:chdp}
A \emph{cooperative hierarchical Dirichlet process (CHDP)} is a distribution over a set of random probability measures (over $\Theta$) located at multiple layers. It defines:
\begin{itemize}
  \item Each layer has with a number $N^{\ell}$ of random probability measures $\{G_i^{\ell}\}_{i=1:N^{\ell}}$ where $N^{1} = 1$ for the first layer;
  \item At the first layer $\ell = 1$, a single global random probability measure $G_0$ is defined, which is distributed as a Dirichlet process parameterized by a concentration parameter $\alpha_0$ and a base probability measure $H$
\begin{equation*}
\begin{aligned}
G_0 \sim DP(\alpha_0, H)
\end{aligned}
\end{equation*}
  \item At the following layer $\ell > 1$, each probability measure $G_i^{\ell}$ at layer $\ell$ is the \emph{Inheritance} from the cooperation of probability measures at the upper layer $\ell-1$ which link to $i$,
\begin{equation*}
\begin{aligned}
G_i^{\ell} \sim DP(\alpha_\ell, G_i^{\ell-1})
\end{aligned}
\end{equation*}
where $\alpha_{\ell}$ is the DP parameter at the layer $\ell$ and $G_i^{\ell-1}$ is from \emph{Superposition} in Definition \ref{def:superposition}
\begin{equation*}
\begin{aligned}
G_i^{\ell-1} &= G_{j_1}^{\ell-1} \oplus G_{j_2}^{\ell-1} \oplus \cdots \oplus G_{J_i}^{\ell-1}\\
\end{aligned}
\end{equation*}
or \emph{Maximization} in Definition \ref{def:maximization}
\begin{equation*}
\begin{aligned}
G_i^{\ell-1} &= G_{j_1}^{\ell-1} \vee G_{j_2}^{\ell-1} \vee \cdots \vee G_{J_i}^{\ell-1}\\
\end{aligned}
\end{equation*}
where each $G_{j}^{\ell-1}$ denotes a random measure at layer $\ell-1$ with a link to $i$ and $\{j_1,\ldots,J_i\}$ are the index of linked measures at layer $\ell-1$.
\end{itemize}
\end{rmk}

The above CHDP has defined a prior, and we should specify the data likelihood to complete the data generation process: to sample a parameter from the bottom layer $\theta_k \sim G^L$ which is used to generate the data $w_{d,n} \sim \theta_k$. $H$ is the base measure of top layer DP and defines the parameter space, which is normally set as a Dirichlet distribution for discrete data (e.g., documents). For example, when applied to \emph{author-document-word}, $\theta_k$ is named the $k$-th topic, $w_{d,n}$ is the $n$-th word of document $d$, and $H$ is a Dirichlet distribution on $(V-1)$-simplex where $V$ is the vocabulary size.

Comparing Definitions \ref{def:hdp} and \ref{def:chdp}, we can draw the conclusion that HDP can be seen as a special case of CHDP with each child node/probability measure having only one parent node/probability measure. If the cooperative/Type-II hierarchical structure degenerates into a Type-I hierarchical structure, the CHDP will degenerate into a HDP as well.

In Figures \ref{fig:hdpforchs} and \ref{fig:chdpforchs}, we compare the graphical models of HDP and CHDP for a particular hierarchical structure: \emph{author-document-word}, where this simple data includes three documents written by two authors and each document $d$ has with $N_d$ words. We also use colors to show how HDP and CHDP are used to model a hieratical structure. It can be seen that the random measures at the author and document layers of the HDP in Figure \ref{fig:hdpforchs} have a one-to-many relationship, where Figure \ref{fig:chdpforchs} (or CHDP) shows a many-to-many relationship. The ability of CHDP to model this many-to-many relationship is due to the designed cooperation. Therefore, CHDP is more powerful than HDP for more general hierarchical structure modeling. Note that the many-to-many relationship between the documents and words are both modeled by HDP and CHDP by the mixture likelihood.

Two similar studies have been published on the convex combination of DPs. Lin and Fisher \cite{lin2012coupling} proposed to use the convex combination of a finite number of DPs $\{G_i^{\ell}\}$ at a high layer as a new measure for the low layer $G^{\ell-1}=\sum_i \omega_i G_i^{\ell}$, and Chen \cite{chen2013dependent} further extended this idea to all normalized random measures with DP as a special case. We want to highlight that although the idea of \emph{Cooperation: Superposition} in this paper is similar to their work, they are different. The idea in \cite{lin2012coupling,chen2013dependent} is to directly use the new measure as the measure of the nodes at a lower layer and the difference between the two new measures relies on the different mixing weights. For example, $G_1^{\ell-1}=\sum_i \omega_{1,i} G_i^{\ell}$ and $G_2^{\ell-1}=\sum_i \omega_{2,i} G_i^{\ell}$ are different only if $\{\omega_{1,i}\}$ are different from $\{\omega_{2,i}\}$. However, in our CHDP, we use this convexly combined measure as the base measure of a new DP which introduces additional flexibility (controlled by $\alpha$) beyond the mixing weights. For example, $G_1^{\ell-1} \sim DP(\alpha, \sum_i \omega_{1,i} G_i^{\ell})$ and $G_2^{\ell-1} \sim DP(\alpha, \sum_i \omega_{2,i} G_i^{\ell})$ may be different even though $\{\omega_{1,i}\}$ and $\{\omega_{2,i}\}$ are the same. When modeling hierarchical structures, it is usually assumed that the whole structure is given and sometimes the mixing weights of the nodes may also be observed. In the situation where mixing weights are known, CHDP shows more model flexibility than the determinate method in \cite{lin2012coupling,chen2013dependent}. Note that we assume the mixing weights are given in this paper and it would be straightforward to model these mixing weights in CHDP just simply adding a Dirichlet prior to them. As for \emph{Cooperation: Maximization}, we found no similar research in the literature.

Next, we introduce two constructive representations for CHDP: international restaurant process representation (marginal one) and stick-breaking representation (explicit one).

\subsection{International restaurant process (IRP) representation}
\label{sec:irp}

The marginal representation of CHDP with $G_0$, $\{G_a\}_{a=1}^A$, and $\{G_d\}_{d=1}^D$ marginalized out (named the international restaurant process) is as follows
\begin{equation}\label{irp:thetaao}
\begin{aligned}
\theta_{a,o} | \theta_{1,1}, \cdots, \theta_{a,o-1}, H &\sim \sum_{k=1}^{K}
\frac{O_{k}}{\sum_k O_k + \alpha_0} \delta_{\theta_{k}} +
\frac{\alpha_0}{\sum_k O_k + \alpha_0} H
\end{aligned}
\end{equation}
\begin{equation}\label{irp:thetadt}
\begin{aligned}
\theta_{d,t} | \theta_{1,1}, \cdots, \theta_{d,t-1}, G_0 &\sim \sum_{o=1}^{O_a}
\frac{T_{a,o}}{\sum_o T_{a,o} + \alpha_a} \delta_{\theta_{a,o}} +
\frac{\alpha_a}{\sum_o T_{a,o} + \alpha_a} G_0
\end{aligned}
\end{equation}
\begin{equation}\label{irp:thetadn}
\begin{aligned}
\theta_{d,n} | \theta_{d,1}, \cdots, \theta_{d,n-1}, G^d_a &\sim
\sum_{t=1}^{T_d}
\frac{N_{d,t}}{\sum_t N_{d,t} + \alpha_d} \delta_{\theta_{d,t}} +
\frac{\alpha_d}{\sum_t N_{d,t} + \alpha_d} G_a^d
\end{aligned}
\end{equation}
where $N_{d,t}$ denotes the number of $\theta_{d,n}$ associated with $\theta_{d,t}$ in $d$; $T_{a,o}$ denotes the number of $\theta_{d,t}$ associated with $\theta_{a,o}$; and $O_k$ denotes the number of $\theta_{a,o}$ associated with $\theta_{k}$. $G_a^d$ is the cooperation between the parent random measures of $d$. If \emph{Superposition} is adopted, then
\begin{equation*}
\begin{aligned}
G_a^d = G_{a_{j_1}} \oplus G_{a_{j_2}} \oplus \cdots \oplus G_{a_{J_d}}
\end{aligned}
\end{equation*}
If \emph{Maximization} is adopted, then
\begin{equation*}
\begin{aligned}
G_a^d = G_{a_{j_1}} \vee G_{a_{j_2}} \vee \cdots \vee G_{a_{J_d}}
\end{aligned}
\end{equation*}
where and $\{a_{j_1}, a_{j_2}, \cdots, a_{J_d}\}$ are authors linked to $d$. The above marginal representation is finished.

Similar to the Chinese restaurant process of DP outlined in Section \ref{sec:dp:crp} and the Chinese restaurant franchise in HDP in Section \ref{sec:hdp:crf}, a metaphor is also introduced to ease the understanding of IRP. Since CHDP is based on a three-layer HDP, we describe the metaphor for the three-layer HDP first, and then introduce one for CHDP. Note that the CRF in Section \ref{sec:hdp:crf} is only a two-layer HDP.

\begin{figure}
\centering
\begin{subfigure}{\textwidth}
\centering
  \includegraphics[scale=0.1]{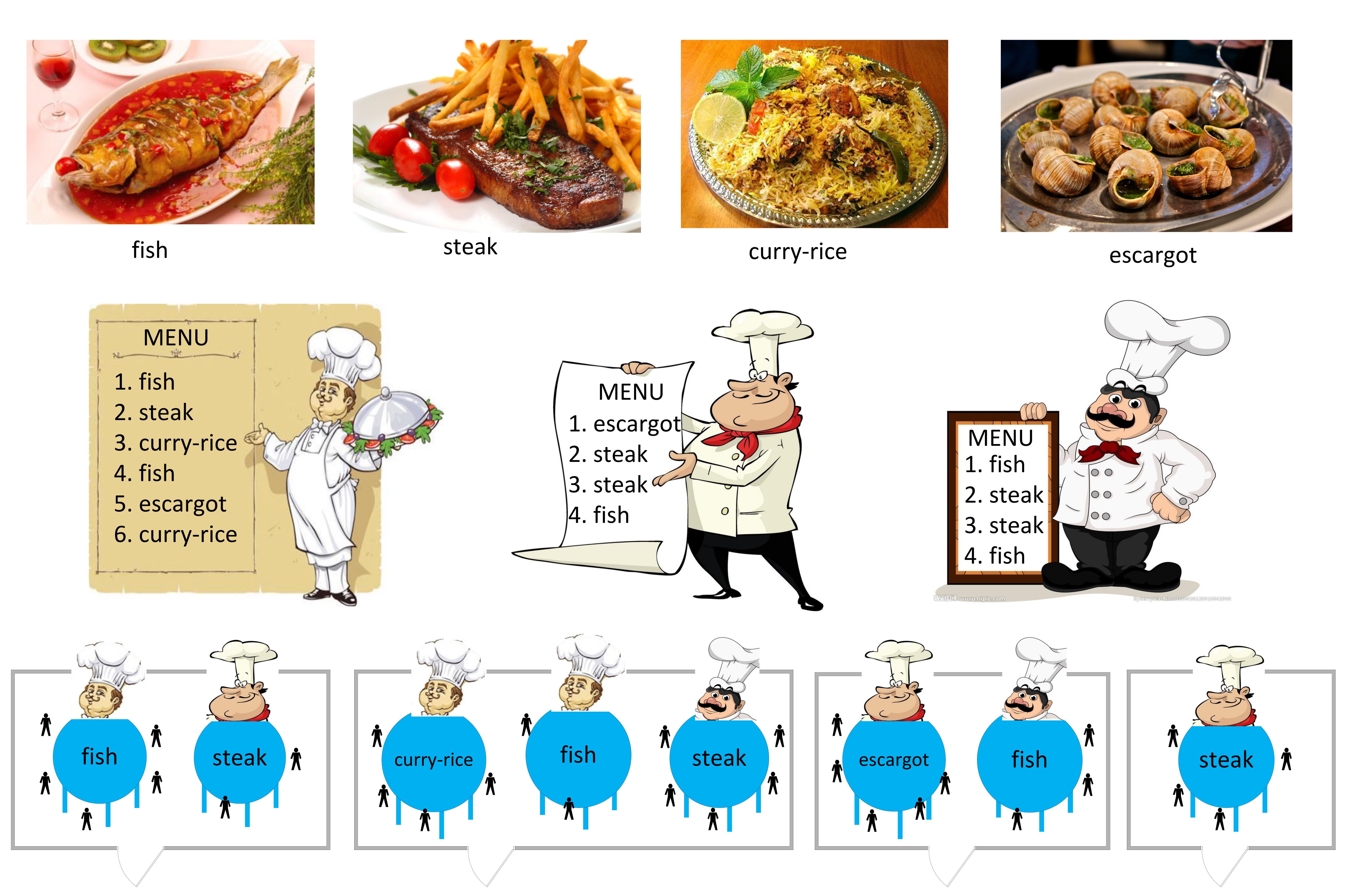}
\caption{International Restaurant Process (IRP)}
\label{irp}
\end{subfigure}
\begin{subfigure}{\textwidth}
\centering
  \includegraphics[scale=0.1]{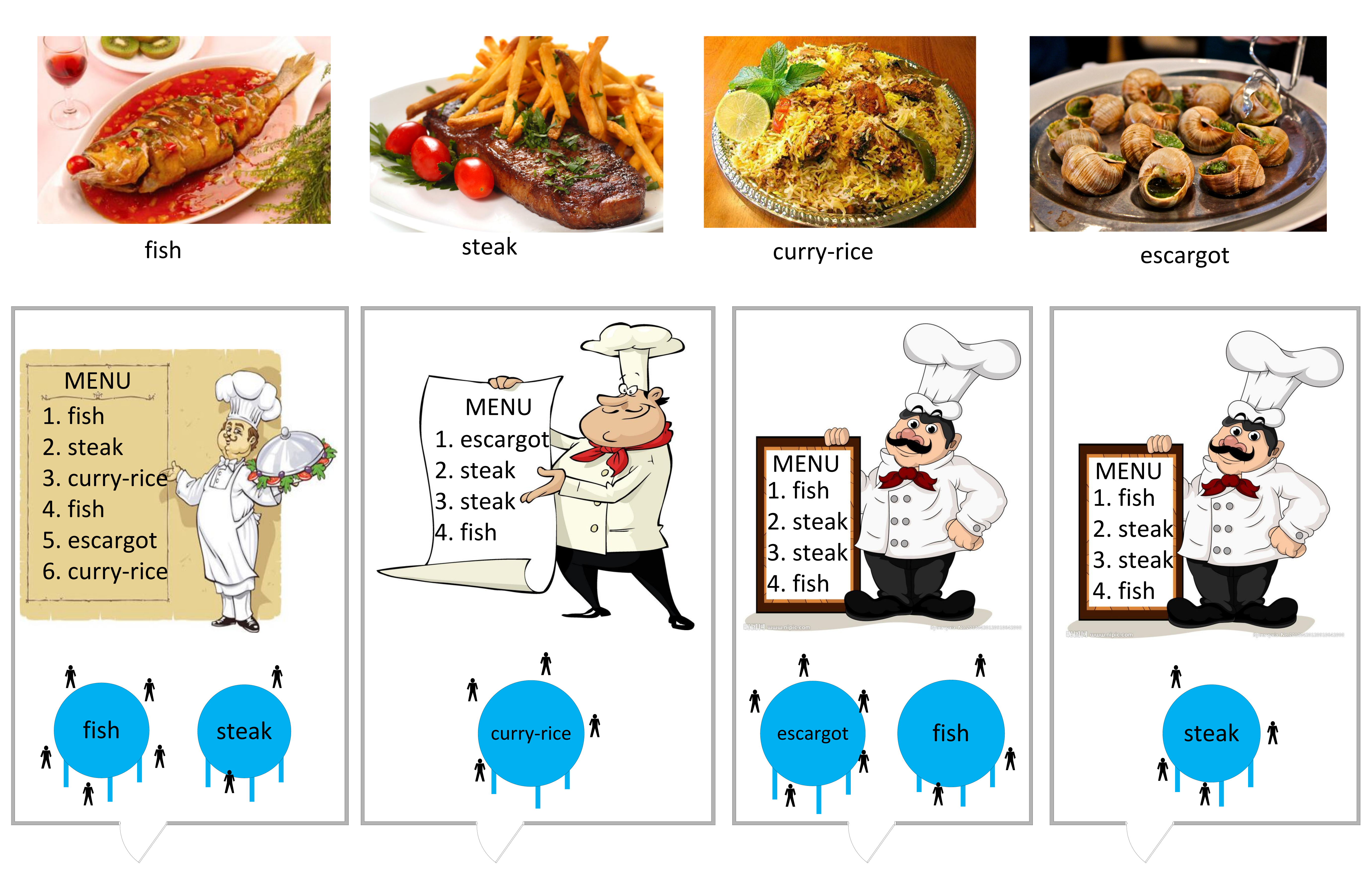}
\caption{(Three-layer) Chinese Restaurant Franchise (CRF)}
\label{crf}
\end{subfigure}
\caption{Comparison of the Chinese restaurant franchise process (three-layer) and the international restaurant process. There are four restaurants and three chefs in the figure. In CRF, all the customers in a restaurant can only be served by one chef, but the customers in IRP can be served by different chefs. The main difference between HDP and CHDP is due to \emph{Cooperation}. }
\label{fig:metphor}
\end{figure}

As shown in Figure \ref{crf}, the metaphor for the three-layer HDP is as follows: there is a global menu with different dishes $\{\theta_k\}_{k=1}^K$ shared by all chefs $\{a\}_{a=1}^A$ from different countries (i.e., China, India, Italy, France). Each chef has a personal menu with dish names as menu options $\{\theta_{a,o}\}$ (Note that menu options are not eliminative - different options could, in fact, be the same dish.) according to their preference and ability. There are also several (national) restaurants $\{d\}$. Each restaurant employs one (and only one) chef, but a chef can work in different restaurants at the same time. For example, a French restaurant hired a French chef, but this chef may work in other French restaurants. In each restaurant, there are multiple tables $T_d$, and each table is served with a dish cooked by the chef of this restaurant. When a customer $n$ walks into a restaurant $d$, she sits at an occupied table with the probability $\frac{N_{d,t}}{\sum_t N_{d,t} + \alpha_d}$ or a new table with the probability $\frac{\alpha_d}{\sum_t N_{d,t} + \alpha_d}$. If an occupied table is selected, she just eats the dish on this table; if the table is new, the customer needs to order a dish for this table from the personal menu of the chef. If option $o$ on the menu is selected with the probability $\frac{T_{a,o}}{\sum_o T_{a,o} + \alpha_a}$, she eats it; if she is not satisfied by all the current options on the menu with the probability $\frac{\alpha_a}{\sum_o T_{a,o} + \alpha_a}$, the chef has to add a new option on the menu from the global shared menu. If dish $k$ on the global menu is selected with the probability $\frac{O_k}{\sum_k O_k + \alpha_0}$, she eats it; if all the dishes on the global menu still do not satisfy this customer with the probability $\frac{\alpha_0}{\sum_k O_k + \alpha_0}$, the chefs have to add a new dish to this global menu (while embarrassedly looking up a recipe book $H$).

As shown in Figure \ref{irp}, the metaphor for the IRP is as follows: the background is almost the same as the one in HDP, but each restaurant in IRP can employ a number of chefs from different countries, and a chef can work in different restaurants. For example, an international restaurant may have a Chinese chef, a French chef, an Italian chef, and an Indian chef (hence its name, \emph{international restaurant}). When a customer $n$ walks into an international restaurant $d$ and needs to order a dish for an empty table, she could order this from the menus of all the chefs working in this restaurant. If option $o$ on the menu of a chef $a$ is selected with the probability $\frac{T_{a,o}}{\sum_o T_{a,o} + \alpha_a}$, she eats it; if she is not satisfied by the current options on the menu with the probability $\frac{\alpha_a}{\sum_o T_{a,o} + \alpha_a}$, she can ask this chef $a$ to add a new option to his menu from the globally shared menu.

\subsection{Stick-breaking representation}
\label{sec:stick}

Based on the stick-breaking process for HDP \cite{teh2006hierarchical}, we develop the following stick-breaking representation for CHDP
\begin{align}
G_0 &= \sum_k^{\infty} \pi_{0,k} \delta_{\theta_k} & \pi_{0,k}&=\nu_{0,k} \prod_{j=1}^{k-1} (1-\nu_{0,j}) & \nu_{0,k} &\sim Beta(1, \alpha_0)
\notag\\
G_a &= \sum_o^{\infty} \pi_{a,o} \delta_{\theta_{a,o}} & \pi_{a,o}&=\nu_{a,o} \prod_{j=1}^{o-1} (1-\nu_{a,j}) & \nu_{a,o} &\sim Beta(1, \alpha_a)
\notag\\
G_d &= \sum_t^{\infty} \pi_{d,t} \delta_{\theta_{d,t}} & \pi_{d,k}&=\nu_{d,t} \prod_{j=1}^{t-1} (1-\nu_{d,j}) & \nu_{d,t} &\sim Beta(1, \alpha_d)
\notag\\
z_{a,o} &\sim \pi_0  &z_{d,t} &\sim \pi_a^d  &  z_{d,n} &\sim \pi_d
\notag\\
w_{d,n} &\sim \theta_{z_{z_{d,z_{d,n}}}}  & & & \theta_k &\sim H
\notag
\end{align}
where $\pi^d_{a}$ is from the cooperation of the parent random measures of $d$:
\begin{itemize}
  \item If \emph{Superposition} is used, then
            \begin{equation}
            \begin{aligned}
            \pi^d_{a} &= \pi_{a_{j_1}} \oplus \pi_{a_{j_2}} \oplus \cdots \oplus \pi_{a_{J_d}}
            \end{aligned}
            \end{equation}
  \item If \emph{Maximization} is used, then
            \begin{equation}
            \begin{aligned}
            \pi^d_{a} &= \pi_{a_{j_1}} \vee \pi_{a_{j_2}} \vee \cdots \vee \pi_{a_{J_d}}
            \end{aligned}
            \end{equation}
\end{itemize}
and $\{a_{j_1}, a_{j_2}, \cdots, a_{J_d}\}$ have links to $d$.
When applied to \emph{author-paper-word}, $\theta_k$ is named the $k$-th topic, $w_{d,n}$ is the $n$-th word of a document $d$, $z_{d,n}$ is the topic assignment of word $n$, and $H$ is a Dirichlet distribution parameterized by $\eta$.

Note that there is no one-to-one mapping between $\pi_{a,o}$ with $\pi_{a,k}$. In fact, their relationship is $\pi_{a,k} = \sum_{o: z_{a,o}=k} \pi_{a,o}$. Similar to $\pi_{d,k}$ and $\pi_{d,t}$, their relation is $\pi_{d,k} = \sum_{t: z_{z_{d,t}}=k} \pi_{d,t}$.

\section{Model Inference}

With the observed CHS, the final aim of the inference is to obtain the posterior distribution of the latent variables in CHDP. Apparently, different representations of CHDP lead to different representations for the posterior distribution. Therefore, we develop one Markov Chain Monte Carlo \cite{ref1} algorithm to approximate the target posterior distribution using samples in Section \ref{CHDP:inf:sampler} based on IRP, and a variational inference \cite{blei2016variational} algorithm to approximate target posterior distribution through optimization in Section \ref{CHDP:inf:vi} based on stick-breaking representation. The main difficulty facing the two inference algorithms lies in cooperation, i.e., \emph{superposition} and \emph{maximization}.

\subsection{Gibbs sampler}
\label{CHDP:inf:sampler}

In this section, we design a Markov Chain Monte Carlo algorithm to obtain samples of the posterior distribution $p(\{\theta_k\}, \{\theta_{a,o}\}, \{\theta_{d,t}\}, K | data, \cdots)$ of CHDP based on IRP representation. Since the difference and difficulty of CHDP comparing three-layer HDP mainly lies on sampling $\theta_{d,t}$, we focus on its inference with two kinds of cooperation: \emph{Superposition} and \emph{Maximization}.

\textbf{Sampling $\theta_{d,t}$ for CHDP-Superposition.} This should be sampled from $G_a^d$, but $G_a^d$ is a superposition of a number of $\{G_a\}$ so it is different from the one in HDP and hard to marginalize out. The $G_a^d$ from \emph{superposition} is,
\begin{equation}
\begin{aligned}
G_a^d \propto \underbrace{\sum_{a_i} \frac{\sum_{o:\theta_{a_i,o}=\theta_1} T_{a_i,o}}{\sum_o T_{a_i,o} + \alpha_a} + \cdots + \sum_{a_i} \frac{\sum_{o:\theta_{a_i,o}=\theta_K}T_{a_i,o}}{\sum_o T_{a_i,o} + \alpha_a}}_{K ~\text{components}}
+ \sum_{a_i} \frac{\alpha_a}{\sum_o T_{a_i,o} + \alpha_a} G_0
\end{aligned}
\end{equation}
where $a_i \in \{a_{j_1}, \ldots, a_{J_d}\}$, the $K$ components of the left-hand side correspond to the observed $K$ dishes, and the remaining part accounts for the new dishes made by the chefs $\{a_{i}\}_{1\le i \le J_d}$. Since \emph{Superposition} is used, each component is a summation across all chefs. Note that the summation also eases the normalization because the summation of the left-hand side is simply $J_d$.

Considering the above $G_a^d$ and IRP representation, $G_a^d$ can be seen as all the menu options of the chefs serving in restaurant $d$, and the sampling of $\theta_{d,t}$ is only a selection procedure from these candidate menu options. Following this idea, we obtain the posterior distribution of $\theta_{d,t}$ as,
\begin{equation}
\label{irps:thetadt}
\begin{aligned}
\theta_{d,t} | \cdots \sim&
\frac{1}{J_d}\sum_{o=1}^{O_{a_{j_1}}} \frac{T_{a_{j_1},o}}{\sum_o T_{a_{j_1},o} + \alpha_a} \delta_{\theta_{a_{j_1},o}} + \frac{1}{J_d}\frac{\alpha_a}{\sum_o T_{a_{j_1},o} + \alpha_a} G_0
\\
&+\cdots
+\frac{1}{J_d}\sum_{o=1}^{O_{a_{J_d}}} \frac{T_{a_{J_d},o}}{\sum_o T_{a_{J_d},o} + \alpha_a} \delta_{\theta_{a_{J_d},o}} + \frac{1}{J_d}\frac{\alpha_a}{\sum_o T_{a_{J_d},o} + \alpha_a} G_0
\end{aligned}
\end{equation}

Another sampling method for CHDP-Superposition is to introduce an auxiliary variable for the sampling of $\theta_{d,n}$ which is given in Appendix 1.

\textbf{Sampling $\theta_{d,t}$ for CHDP-Maximization.} Similar to CHDP-Superposition, the difficulty also lies in the fact that the $G_a^d$ is a maximization of a number of $\{G_a\}$ here. The $G_a^d$ from \emph{maximization} is,
\begin{equation}
\begin{aligned}
G_a^d \propto \underbrace{\max_{a_i} \frac{\sum_{o:\theta_{a_i,o}=\theta_1}T_{a_i,o}}{\sum_o T_{a_i,o} + \alpha_a} + \cdots +
\max_{a_i} \frac{\sum_{o:\theta_{a_i,o}=\theta_K} T_{a_i,o}}{\sum_o T_{a_i,o} + \alpha_a}}_{K ~\text{components}}
+ \sum_{a_i} \frac{\alpha_a}{\sum_o T_{a_i,o} + \alpha_a} G_0
\end{aligned}
\end{equation}
Under IRP representation, the sampling $\theta_{d,t}$ here could also be considered as a menu option selecting procedure. Compared with CHDP-Superposition, the difference is that not all the menu options of chefs serving in restaurant $d$ are seen as candidates. CHDP-Maximization only takes the menu options from the chefs who are the best at these options as the candidates.
Finally, the posterior distribution of $\theta_{d,t}$ is,
\begin{equation}
\label{irpm:thetadt}
\begin{aligned}
\theta_{d,t} \sim&
\sum_{o=1}^{O_{a_{j_1}}} \frac{T_{a_{j_1},o}}{\sum_o T_{a_{j_1},o} + \alpha_a} \mathds{1}\left(a_{j_1} = \arg\max_{a_i} \frac{\sum_{o:\theta_{a_i,o}=\theta_{a_{j_1},o}}T_{a_i,o}}{\sum_o T_{a_i,o} + \alpha_a}\right)\delta_{\theta_{a_{j_1},o}}
\\
&+ \frac{\alpha_a}{\sum_o T_{a_{j_1},o} + \alpha_a} G_0
\\
&+\cdots
+\sum_{o=1}^{O_{a_{J_d}}} \frac{T_{a_{J_d},o}}{\sum_o T_{a_{J_d},o} + \alpha_a}\mathds{1}\left(a_{J_d} = \arg\max_{a_i} \frac{\sum_{o:\theta_{a_i,o}=\theta_{a_{J_d},o}}T_{a_i,o}}{\sum_o T_{a_i,o} + \alpha_a}\right) \delta_{\theta_{a_{J_d},o}}
\\
&+ \frac{\alpha_a}{\sum_o T_{a_{J_d},o} + \alpha_a} G_0
\end{aligned}
\end{equation}
where $\mathds{1}()$ is the identity function which is equal to 1 if the condition is satisfied; 0, otherwise. Here, the identity functions serve as the candidate filter. Note that the normalization is nontrivial for CHDP-Maximization because some options are removed from the candidate list and then the unit summation for each chef does not hold any more.

The posterior distributions of the remaining variables simply follow the three-layer HDP. Due to the space limitation, we list the distributions of the remaining variables in Appendix 1. The entire procedure for the inference of IRP is summarized in Algorithm \ref{ag:irp}.

\subsection{Variational inference}
\label{CHDP:inf:vi}

Different from the designed sampler in the previous section which uses samples to approximate the posterior distribution of latent variables, variational inference \cite{blei2016variational} casts this distribution approximation problem to an optimization problem. While samplers have the advantage of asymptotically exact, they are usually not efficient in practice when facing large-scale data. Optimization-based variational inference \cite{wang2011online} is more tractable than samplers with only a small loss in terms of theoretical accuracy. We therefore develop a variational inference algorithm for CHDP, described as follows, to handle large-scale data.

The core idea of variational inference is to propose a number of (normally independent) variational distributions of latent variables with corresponding variational parameters and to reduce the distance (usually Kullback-Leibler (KL) divergence) between the real posterior distribution and these variational distributions through adjusting the value of these variational parameters. However, the infinite number of factors and their weights make the posterior inference of the stick weights even harder. One common work-around in nonparametric Bayesian learning is to use a truncation method. The truncation method \cite{fox2009bayesian,willsky2009nonparametric}, which uses a relatively big $K^{\dag}$ as the (potential) maximum number of topics, is widely accepted. For CHDP, we define the following variational distributions for the latent variables using stick-breaking representation:
\begin{equation*}
\begin{aligned}
q(\nu_{0,k}) &= \prod_{k=1}^{K^{\dagger}-1}q(\nu_{0,k}; u_{0,k}, r_{0,k})
~~~
&q(\nu_{a}) &= \prod_{a=1}^{A}\prod_{o=1}^{O^{\dagger}-1}q(\nu_{a,o}; u_{a,o}, r_{a,o})
\\
q(\nu_{d}) &= \prod_{d=1}^{D}\prod_{t=1}^{T^{\dagger}-1}q(\nu_{d,t}; u_{d,t}, r_{d,t})
~~~
&q(z_{a,o}) &= \prod_{a=1}^{A}\prod_{o=1}^{O_a}q(z_{a,o}; \varsigma_{a,o})
\\
q(z_{d,t}) &= \prod_{d=1}^{D}\prod_{t=1}^{T_d}q(z_{d,t}; \varsigma_{d,t})
~~~
&q(z_{d,n}) &= \prod_{d=1}^{D}\prod_{n=1}^{N_d}q(z_{d,n}; \varsigma_{d,n})
\\
q(\theta) &= \prod_{k=1}^{K^{\dagger}}q(\theta_k; \vartheta_k)&&
\end{aligned}
\end{equation*}
where $H$ is chosen as $Dir(\eta)$, $K^{\dagger}$, $O^{\dagger}$ and $T^{\dagger}$ are the truncation levels, $\nu_{0, K^{\dagger}} = 1$, $\{\nu_{a, O^{\dagger}} = 1\}$, $\{\nu_{d, T^{\dagger}} = 1\}$, $\{u, r, \varsigma, \vartheta\}$ are the defined variational parameters. With these variational distributions, we have
\begin{equation*}
\begin{aligned}
&\log p(w|\alpha_0, \alpha_a, \alpha_d, \eta)
\\
\ge&  \mathbb{E}_q \left [ \log p(w, \nu_0, \nu_a, \nu_d, z_{a,o}, z_{d,t}, z_{d,n}, \theta |\alpha_0, \alpha_a, \alpha_d, \eta) \right ]
\\
&- \mathbb{E}_q \left [ \log q(\nu_0, \nu_a, \nu_d, z_{a,o}, z_{d,t}, z_{d,n}, \theta) \right ]
\\
=& \pounds(q)
\\
=&\log p(w|\alpha_0, \alpha_a, \alpha_d, \eta)
\\
&- \mathbb{D}_{KL}[q(\nu_0, \nu_a, \nu_d, z_{a,o}, z_{d,t}, z_{d,n}, \theta) || p(\nu_0, \nu_a, \nu_d, z_{a,o}, z_{d,t}, z_{d,n}, \theta |w, \alpha_0, \alpha_a, \alpha_d, \eta)]
\end{aligned}
\end{equation*}
where $\pounds(q)$ is the evidence lower bound (ELBO). Our objective is to maximize ELBO through updating variational parameters, and maximizing of ELBO is equal to minimizing the KL divergence between the real posterior distribution and the variational distribution. Next, we use the coordinate gradient optimization method to update the variational parameters.

\textbf{Update $\varsigma_{a,o,k}$ for CHDP-Superposition.} The derivative of $\pounds(q)$ with respective to $\varsigma_{a,o,k}$ is
\begin{align*}
\frac{\partial \pounds_{\varsigma_{a,o,k}}(q)}{\partial \varsigma_{a,o,k}}
=&\left (\Psi(u_{0,k}) - \Psi(u_{0,k} + r_{0,k}) \right )+\sum_{h<k}\left(\Psi(r_{0,h}) - \Psi(u_{0,h} + r_{0,h})\right)
\\
&+\sum_d \sum_n \sum_t \varsigma_{d,t,ao}  \sum_v \varsigma_{d,n,t} \delta(w_{d,n}=v)
\left ( \Psi(\vartheta_{k,v}) - \Psi\left(\sum_v \vartheta_{k,v}\right) \right )
\\
& -\log \varsigma_{a,o,k} - 1
\end{align*}
Note that updating $\varsigma_{a,o,k}$ using this derivative with a step $\tau$ implies a Euclidean regularization $\frac{1}{2\tau}||\varsigma_{a,o,k} - \varsigma^{(i)}_{a,o,k}||^2$ where $\varsigma_{a,o,k}^{(i)}$ is the value in the last ($i$-th) iteration. This update overlooks the geometry of the variable, i.e., the changes of a variational distribution and its variational parameters are not synchronous. To consider the distribution geometry, natural gradient \cite{HoffmanBWP13} and proximal gradient methods \cite{KhanBFF15} are proposed in the literature. Here, we adopt the proximal gradient method to resolve our problem, which has better convergence properties \cite{KhanBLSS16}. Note that all the following variational parameter updates use proximal regularization. For $\varsigma_{a,o,k}$, we introduce an additional regularization $-\gamma \mathbb{D}_{KL}[q(z_{a,o} | \varsigma_{a,o,k})||q( z_{a,o}| \varsigma^{(i)}_{a,o,k})]$, and then the new derivative becomes
\begin{align*}
\frac{\partial \pounds_{\varsigma_{a,o,k}}(q)}{\partial \varsigma_{a,o,k}}
=&\left (\Psi(u_{0,k}) - \Psi(u_{0,k} + r_{0,k}) \right )+\sum_{h<k}\left(\Psi(r_{0,h}) - \Psi(u_{0,h} + r_{0,h})\right)
\\
&+\sum_d \sum_n \sum_t \varsigma_{d,t,ao}  \sum_v \varsigma_{d,n,t} \delta(w_{d,n}=v)
\left ( \Psi(\vartheta_{k,v}) - \Psi\left(\sum_v \vartheta_{k,v}\right) \right )
\\
&-(1+\gamma)\log \varsigma_{a,o,k} - (1+\gamma) + \gamma \log\varsigma^{(i)}_{a,o,k}
\end{align*}
Finally, it can be updated by
\begin{equation}
\label{vis:varsigmaaok}
\begin{aligned}
\varsigma_{a,o,k}^{(i+1)} \propto &\exp \Bigg\{
\frac{1}{1+\gamma}
\Bigg (
\left (\Psi(u_{0,k}) - \Psi(u_{0,k} + r_{0,k}) \right )+\sum_{h<k}\left(\Psi(r_{0,h}) - \Psi(u_{0,h} + r_{0,h})\right)
\\
&~~~~~~~~~~~~~~~~- (1+\gamma) +\gamma\log\varsigma^{(i)}_{a,o,k}
\\
&~~~~~~~~~~~~~~~~
+\sum_d \sum_n \sum_t \varsigma_{d,t,ao}  \sum_v \varsigma_{d,n,t} \delta(w_{d,n}=v)
\left ( \Psi(\vartheta_{k,v}) - \Psi\left(\sum_v \vartheta_{k,v}\right) \right )
\Bigg )\Bigg\}
\end{aligned}
\end{equation}
Note that when updating $\varsigma_{a,o,K}$, the item, i.e., $\Psi(u_{0,k}) - \Psi(u_{0,k} + r_{0,k})$ should be removed because $\nu_{0, K}=1$.

\textbf{Update $\varsigma_{d,t,ao}$ for CHDP-Superposition.} The EBLO with $\varsigma_{d,t,ao}$ is
\begin{align*}
\pounds_{\varsigma_{d,t}}(q) =&
\mathbb{E}_q \left [ \sum_d \sum_t \log p(z_{d,t}|\{\nu_{a}\}) \right ]
+ \mathbb{E}_q \left [ \sum_d \sum_n  \log p(w_{d,n}| \theta, z_{a,o}, z_{d,t}, z_{d,n}) \right ]
\\
&- \mathbb{E}_q \left [ \sum_d \sum_t \log q(z_{d,t}| \varsigma_{d,t})\right ]
\end{align*}
where
\begin{align*}
p(z_{d,t}|\{\nu_{a}\}) = \prod_{ao} \left(\pi_{ao}^d \right)^{\delta(z_{d,t}=ao)},~~~\pi_{ao}^d = \frac{\pi_{a,o}}{J_d}
\end{align*}
and then
\begin{equation}
\begin{aligned}
& \mathbb{E}_q \left [ \sum_d \sum_t \log p(z_{d,t}|\{\nu_{a}\}) \right ]
\\
=& \sum_d \sum_t \sum_{a \in a_d} \sum_{o \in a} \mathbb{E}_q \left [ \log \left(\frac{\pi_{a,o}}{J_d}\right)^{\delta(z_{d,t}=ao)} \right ]
\\
=& \sum_d \sum_t \sum_{a \in a_d} \sum_{o \in a}
\varsigma_{d,t,ao}\left(\left (\Psi(u_{a,o}) - \Psi(u_{a,o} + r_{a,o}) \right ) +\sum_{h<o, h\in a} \left (\Psi(r_{a,h}) - \Psi(u_{a,h} + r_{a,h}) \right ) - \log J_d \right)
\end{aligned}
\label{expzdt}
\end{equation}
The above result is relatively simple, because the normalization in \emph{Superposition} is intuitive: normalizing of $\{\pi_{a,o}\}$ is done simply by multiplying $\frac{1}{J_d}$ because $\sum_o \pi_{a,o} = 1$ and $\sum_a\sum_o \pi_{a,o} = J_d$ thanks to the linearity nature of \emph{Superposition}. This simplicity does not hold for \emph{Maximization} where normalizing $\pi_{a,o}$ depends on other $\{\pi_{\bar{a},\bar{o}}| \bar{a} \neq a, \bar{o} \neq o\}$, which will be discussed in more detail in its update for CHDP-Maximization.

Finally, it can be updated by
\begin{equation}
\begin{aligned}
\varsigma_{d,t,ao}^{(i+1)} \propto &\exp \Bigg\{
\frac{1}{1+\gamma}
\Bigg (
\left(\left (\Psi(u_{a,o}) - \Psi(u_{a,o} + r_{a,o}) \right ) +\sum_{h<o, h\in a} \left (\Psi(r_{a,h}) - \Psi(u_{a,h} + r_{a,h}) \right ) - \log J_d \right)
\\
&~~~
- (1+\gamma) +\gamma\log\varsigma^{(i)}_{d,t,ao}
\\
&~~~+\sum_n \sum_k \varsigma_{a,o,k} \sum_v \varsigma_{d,n,t} \delta(w_{d,n}=v)
\left ( \Psi(\vartheta_{k,v}) - \Psi\left(\sum_v \vartheta_{k,v}\right) \right )
\Bigg )\Bigg\}
\end{aligned}
\label{vis:varsigmadtao}
\end{equation}
Similarly, when updating $\varsigma_{d,t,aO}$, the item, i.e., $\Psi(u_{a,o}) - \Psi(u_{a,o} + r_{a,o})$ should be removed because $\nu_{a, O}=1$.

\textbf{Update $u_{a,o}$ and $r_{a,o}$ for CHDP-Superposition.} Ignoring the detailed deduction, they can be updated by
\begin{equation}
\begin{aligned}
u^{(i+1)}_{a,o} = \frac{\sum_d \sum_t\varsigma_{d,t,ao} + \gamma( u^{(i)}_{a,o}-1)}{1+\gamma}+1
\end{aligned}
\label{vis:uao}
\end{equation}
and
\begin{equation}
\begin{aligned}
r^{(i+1)}_{a,o} = \frac{\alpha_a -1 + \sum_d \sum_t \sum_{h>o,h\in a} \varsigma_{d,t,ah}  +\gamma(r^{(i)}_{a,o}-1)}{1+\gamma} +1
\end{aligned}
\label{vis:rao}
\end{equation}

\textbf{Update $\varsigma_{a,o,k}$ for CHDP-Maximization.} The ELBO with respective to $\varsigma_{a,o,k}$ is,
\begin{align*}
\pounds_{\varsigma_{a,o,k}}(q) =&
\mathbb{E}_q \left [ \sum_a \sum_o \log p(z_{a,o}|\nu_{0}) \right ]
+ \mathbb{E}_q \left [ \sum_d \sum_t \log p(z_{d,t}|\{\nu_{a}\}, \{z_{a,o}\}) \right ]
\\
&+ \mathbb{E}_q \left [ \sum_d \sum_n  \log p(w_{d,n}| \theta, z_{a,o}, z_{d,t}, z_{d,n}) \right ]
- \mathbb{E}_q \left [ \sum_a \sum_o \log q(z_{a,o}| \varsigma_{a,o})\right ]
\end{align*}
where
\begin{equation}
\label{vi:pzdt}
\begin{aligned}
p(z_{d,t}|\{\nu_{a}\}, \{z_{a,o}\}) = \prod_{ao} \left(\pi_{ao}^d \right)^{\delta(z_{d,t}=ao)}
\end{aligned}
\end{equation}
and
\begin{align}
\label{vi:piaod}
\pi_{ao}^d = \frac
{\pi_{a,o} \mathds{1} \left(a=\arg\max\limits_{a_i} \left \{ \sum_{\{o: z_{a_{j_1},o}=z_{a,o}\}} \pi_{a_{j_1},o},\cdots, \sum_{\{o:z_{a_{J_d},o}=z_{a,o}\}} \pi_{a_{J_d},o} \right \}\right)}
{\sum_{ao} \pi_{a,o}\delta \left(a=\arg\max\limits_{a_i} \left \{ \sum_{\{o: z_{a_{j_1},o}=z_{a,o}\}} \pi_{a_{j_1},o},\cdots, \sum_{\{o: z_{a_{J_d},o}=z_{a,o}\}} \pi_{a_{J_d},o} \right \}\right)}
\end{align}
Comparing the update for CHDP-Superposition, there is an additional item (i.e., the second expectation) in $\pounds_{\varsigma_{a,o,k}}(q)$. The reason is that $z_{d,t}$ is independent with $z_{a,o}$ in CHDP-Superposition but $z_{d,t}$ depends on $z_{a,o}$ in CHDP-Maximization according to the aforementioned probability of $\pi^d_{ao}$. Due to the complicated functional form of this probability, it is difficult to evaluate this expectation and obtain its derivative with a closed form. Next, we try to approximate this expectation and its derivative,
\begin{align*}
&\mathbb{E}_q \left [ \sum_d \sum_t \log p(z_{d,t}|\{\nu_{a}\}, \{z_{a,o}\}) \right ]
\\
\approx& \sum_d \sum_t \sum_{a \in a_d} \sum_{o \in a}
\varsigma_{d,t,ao}
\left(\varsigma_{a,o,k} \nabla_{\varsigma_{a,o,k}}  \mathbb{E}_q \left [\log \pi_{ao}^d \right ]  \right)
\\
=& \sum_d \sum_t \sum_{a \in a_d} \sum_{o \in a}
\varsigma_{d,t,ao}
\left(\varsigma_{a,o,k}  \mathbb{E}_q \left [\log \pi_{ao}^d \nabla \log q(z_{a,o}|\varsigma_{a,o,k})\right ]  \right)
\\
=& \sum_d \sum_t \sum_{a \in a_d} \sum_{o \in a}
\varsigma_{d,t,ao}
\left(\varsigma_{a,o,k}  \mathbb{E}_q \left [
\frac{\log \pi_{ao}^d \delta(z_{a,o}=k)}{\varsigma^{(i)}_{a,o,k}}\right ]  \right)
\\
\approx& \sum_d \sum_t \sum_{a \in a_d} \sum_{o \in a}
\varsigma_{d,t,ao}
\left(\varsigma_{a,o,k}
\frac{1}{S}\sum_s
\frac{\log (\pi_{ao}^d)^{(s)} \delta(z^{(s)}_{a,o}=k)}{\varsigma^{{i}}_{a,o,k}}  \right)
\end{align*}
where $\log (\pi_{ao}^d)^{(s)}$ is evaluated by replacing $z_{a}$ and $\pi_{a}$ in Eq. (\ref{vi:piaod}) by a set of samples $z^{(s)}_{a}$ and $\pi^{(s)}_{a}$. The first approximation holds due to linear approximation that is also adopted by the Laplace variational inference \cite{Wang2013vin} and proximal variational inference \cite{KhanBFF15} for the non-conjugate situation; the second equality holds with the help of the score function estimator \cite{SchulmanHWA15} which is used to move the derivative into the expectation and avoid computing the derivative of $\Omega$; the last approximation is done through Monte Carl, using $S$ samples from
$\prod_a\prod_o q(\nu_{a,o}| u^{(i)}_{a,o}, r^{(i)}_{a,o}) q(z_{a,o}|\varsigma^{(i)}_{a,o,k})$ to approximate the expectation. Finally, we obtain an unbiased stochastic estimate of the derivative (with proximal regularization) as
\begin{equation}
\label{vim:varsigmaaok}
\begin{aligned}
\frac{\partial \pounds_{\varsigma_{a,o,k}}(q)}{\partial \varsigma_{a,o,k}}
=&\left (\Psi(u_{0,k}) - \Psi(u_{0,k} + r_{0,k}) \right )+\sum_{h<k}\left(\Psi(r_{0,h}) - \Psi(u_{0,h} + r_{0,h})\right)
\\
&+ \sum_d \sum_t \sum_{a \in a_d} \sum_{o \in a}
\varsigma_{d,t,ao}
\left(
\frac{1}{S}\sum_s
\frac{\log (\pi_{ao}^d)^{(s)}\delta(z^{(s)}_{a,o}=k)}{\varsigma^{{i}}_{a,o,k}}  \right)
\\
&+\sum_d \sum_n \sum_t \varsigma_{d,t,ao}  \sum_v \varsigma_{d,n,t} \delta(w_{d,n}=v)
\left ( \Psi(\vartheta_{k,v}) - \Psi\left(\sum_v \vartheta_{k,v}\right) \right )
\\
&-(1+\gamma)\log \varsigma_{a,o,k} - (1+\gamma) + \gamma \log\varsigma^{(i)}_{a,o,k}
\end{aligned}
\end{equation}
Finally, it can be updated by a step towards its gradient.

\textbf{Update $\varsigma_{d, t, ao}$ for CHDP-Maximization.} The update equation contains an expectation of $\log \pi_{ao}^d$ which is again approximated by the Monte Carlo. Ignoring the detailed deduction, we can obtain its derivative as follow
\begin{equation}
\label{vim:varsigmadtao}
\begin{aligned}
\varsigma_{d,t,ao}^{(i+1)} \propto &\exp \Bigg\{
\frac{1}{1+\gamma}
\Bigg (
\frac{1}{S}\sum_s \log (\pi_{ao}^d)^{(s)}
- (1+\gamma) +\gamma\log\varsigma^{(i)}_{d,t,ao}
\\
&~~~
+\sum_n \sum_k \varsigma_{a,o,k} \sum_v \varsigma_{d,n,t} \delta(w_{d,n}=v)
\left ( \Psi(\vartheta_{k,v}) - \Psi\left(\sum_v \vartheta_{k,v}\right) \right )
\Bigg )\Bigg\}
\end{aligned}
\end{equation}

\textbf{Update $u_{a,o}$ and $r_{a,o}$ for CHDP-Maximization.} The update of variational parameters $u_{0,k}$ and $r_{0,k}$ also encounters problem in the update of $\varsigma_{a,o,k}$ as it is difficult to obtain the derivative of a complicated expectation. Again, we use the same strategies for updating $\varsigma_{a,o,k}$ for CHDP-Maximization. Ignoring the detailed deductive, the final derivatives are
\begin{align}
\label{vim:uao}
\frac{\partial \pounds_{u_a}(q)}{\partial u_{a,o}}
=&\left(-\Psi^{\prime}(u_{a,o} + r_{a,o})\right)
\Bigg (\alpha_a-1 - (1+\gamma)(u_{a,o}-1) -(1+\gamma)(r_{a,o} -1)
\nonumber\\
& + \gamma(u^{(i)}_{a,o}-1) + \gamma(r_{a,o} -1)
\Bigg)
\nonumber\\
& + \Psi^{\prime}(u_{a,o}) \Bigg (
\gamma( u^{(i)}_{a,o}-1) -(1+\gamma)(u_{a,o}-1)
\Bigg)
\\
&+ \sum_d \sum_t
\varsigma_{d,t,ao} \Bigg( \frac{1}{S}\sum_s \log (\pi_{ao}^d)^{(s)}
(\Psi(u^{(i)}_{a,o}+r_{a,o}) - \Psi(u^{(i)}_{a,o}) + \log\nu^{(s)}_{a,o} )
\Bigg)\nonumber
\end{align}
and
\begin{align}
\label{vim:rao}
\frac{\partial \pounds_{r_a}(q)}{\partial r_{a,o}}
=& (- \Psi^{\prime}(u_{a,o} + r_{a,o}))
\Bigg(\alpha_a-1
- (1+\gamma)(r_{a,o}-1) - (1+\gamma)(u_{a,o}-1)
\nonumber\\
&+\gamma(r^{(i)}_{a,o}-1) +\gamma(u_{a,o}-1)
\Bigg)
\nonumber\\
&+\Psi^{\prime}(r_{a,o})
\Bigg ( \alpha_a-1 - (1+\gamma)(r_{a,o}-1) +\gamma(r^{(i)}_{0,k}-1) \Bigg )
\\
&+\sum_d \sum_t
\varsigma_{d,t,ao} \Bigg(\frac{1}{S}\sum_s  \log (\pi_{ao}^d)^{(s)}
(\Psi(u_{a,o}+r^{(i)}_{a,o}) - \Psi(r^{(i)}_{a,o}) + \log(1-\nu^{(s)}_{a,o}) )
 \Bigg)\nonumber
\end{align}

Since the update of the remaining variational parameters, e.g.,  $u_{0,k}$ and $r_{0,k}$, are common for both CHDP-Superposition and CHDP-Maximization and relatively simple, they are given in Appendix 2 to complete the entire procedure. Finally, the whole variational inference algorithm is summarized in Algorithm \ref{ag:vi}. Note that this algorithm is demonstrated for three-layer hierarchical structure modeling. It is interesting that Algorithm \ref{ag:vi} is an alternative sampling and optimizing procedure, e.g., the update of variational parameters at layer $A$ needs the samples of the latent variables at this layer in advance. When applying this on the hierarchical structure with more than three layers, the update of variational parameters (e.g., $u$ and $r$) for each layer will need samples of the latent variable at this layer.

\section{Experiments}

We present experimental evaluations of the proposed CHDP regarding its properties and practical usefulness. We first present a set of experiments on synthetic data to analyze the properties of CHDP and the designed inference algorithms, i.e., the convergence analysis of the proposed MCMC algorithms (in Section \ref{sec:convergence}), the parameter sensitivity analysis of CHDP (in Section \ref{sec:parameters}), and the ability to uncover the hidden structure comparing its base model: HDP (in Section \ref{sec:cooperative}). We then move to the real-world setting, where we evaluate the performance of CHDP on two real-world applications based on real-world datasets compare with state-of-the-art models or algorithms on these applications (in Section \ref{sec:realworld}).

\subsection{Evaluation on the convergence of the designed samplers}
\label{sec:convergence}

\begin{figure}[!t]
\centering
\includegraphics[scale=0.45]{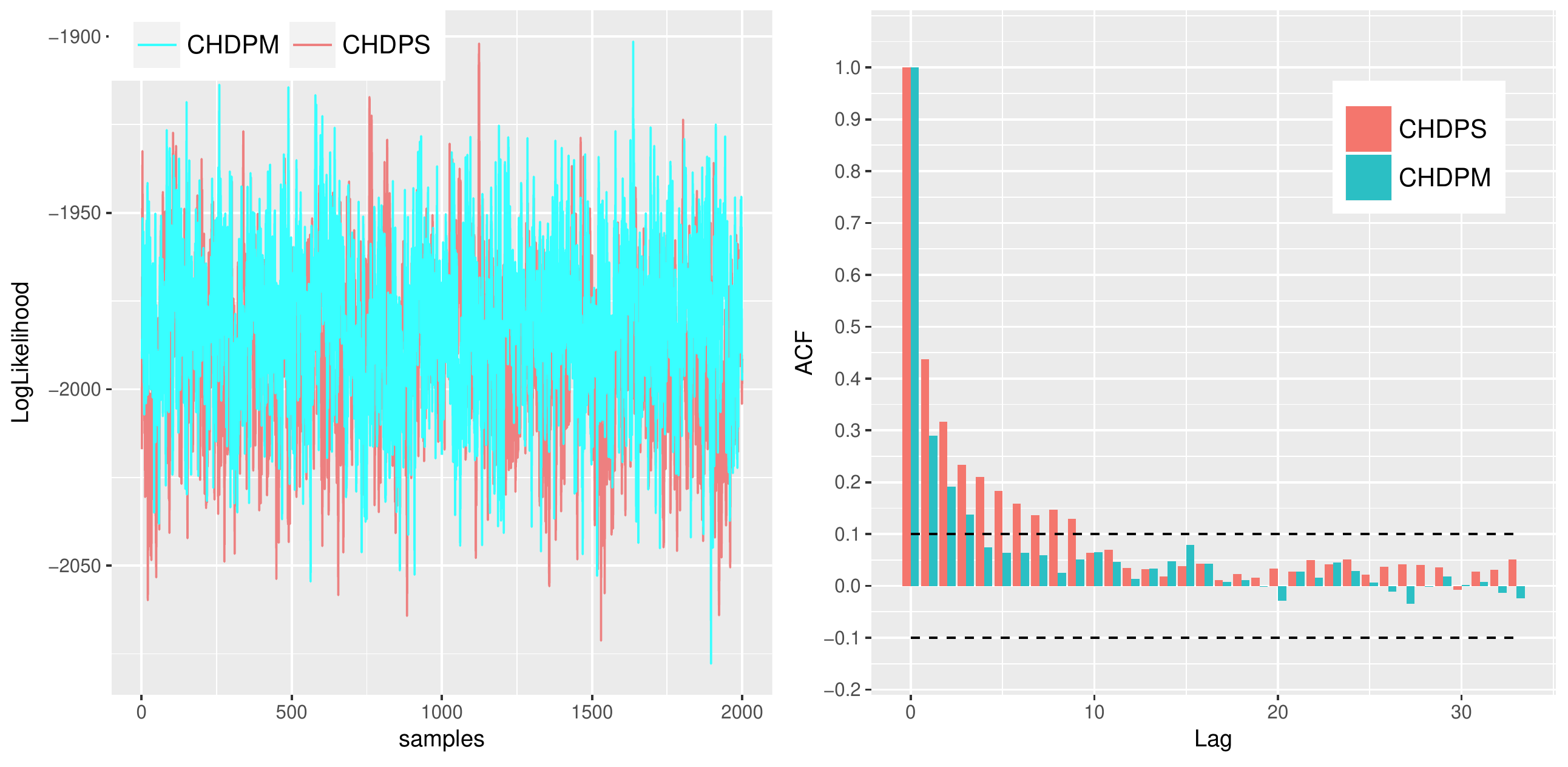}
\caption{Convergence comparison between CHDPS and CHDPM using one chain on \emph{LogLikelihood} and \emph{ACF}. (The sample number is 2,000, and it is acceptable that the chain is convergent if ACF is smaller than 0.1.)}
\label{chdp:cvgL}
\end{figure}

\begin{figure}[!t]
\centering
\includegraphics[scale=0.45]{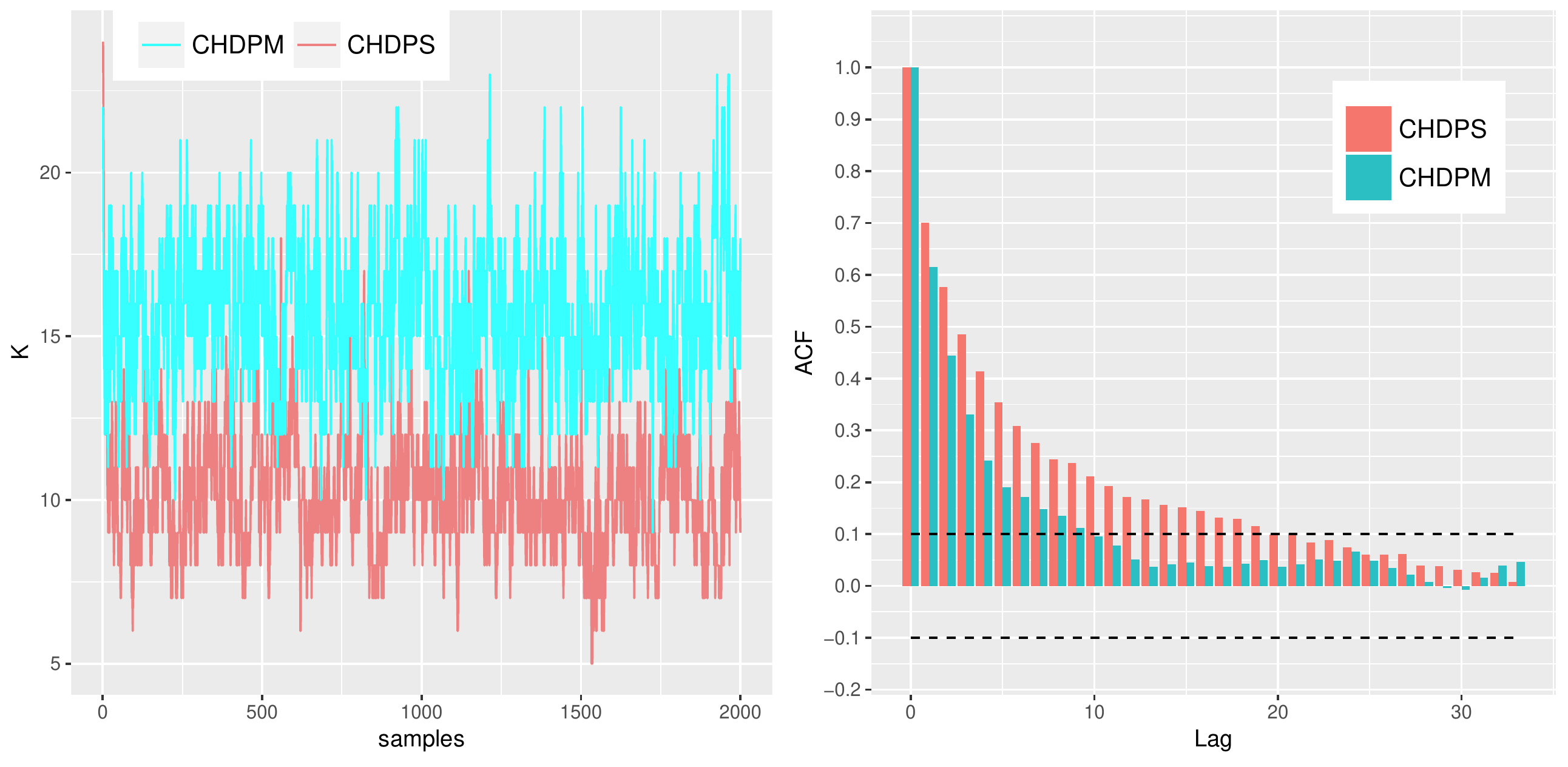}
\caption{Convergence comparison between CHDPS and CHDPM using one chain on \emph{K} and \emph{ACF}. (Sample number is 2,000, and it is acceptable that the sampler is convergent if ACF is smaller than 0.1.)}
\label{chdp:cvgK}
\end{figure}

In Section 5, we presented two inference algorithms for CHDP: MCMC-based and Optimization-based. Optimization-based inference algorithms can easily track its convergence through evaluating the ELBO, but it is not easy to assess the convergence of MCMC-based inference algorithms \cite{brooks1998assessing}. Therefore, we need to evaluate the convergence of the designed samplers (Algorithm \ref{ag:irp}) for CHDPS and CHDPM. In the literature, the methods for the convergence analysis of MCMC are roughly grouped into two categories: one chain-based or multiple (normally 3 to 7) chains-based. We first randomly generated a hierarchical structure with $A=20,D=50,V=100$: each document had 10 words, the links between authors and documents were randomly generated, and the mixing density was 0.3 with a guarantee that each author linked to at least one document and each document had at least one author, the model parameters were $\alpha_0=1, \alpha_a=1, \alpha_d=1, \eta=0.5$. On this synthetic data, we ran both CHDPS and CHDPM and collected 2,000 samples, and then Autocorrelation (ACF) \cite{geyer1992} was used for the convergence evaluation of CHDPS and CHDPM based on their chains. In Fig. \ref{chdp:cvgL}, the \emph{Loglikelihood} of two samplers was plotted along samples on the left-hand side, and the evaluated ACF values were plotted along different lags on right-hand side. In Fig. \ref{chdp:cvgK}, the hidden factor number \emph{K} of two samplers was plotted along samples on the left-hand side, and the evaluated ACF values were plotted along different Lags on right-hand side. Furthermore, we also plotted two dashed lines with ACF values $0.1$ and $-0.1$ on the right-hand side in both two figures, because a sampler is believed converge well if its ACF absolute value is smaller than 0.1. The reason why \emph{Loglikelihood} and \emph{K} are selected as the representatives of two samplers is that they are highly dependent on all the latent variables and if they are convergent, other latent variables will also be convergent. According to two figures, we can draw the following conclusions: 1) two models can converge well because the ACF values were finally smaller than 0.1; 2) \emph{Loglikelihood} converged more quickly than \emph{K}; 3) CHDPM converged more quickly than CHDPS.

\begin{figure}[!t]
\centering
\includegraphics[scale=0.52]{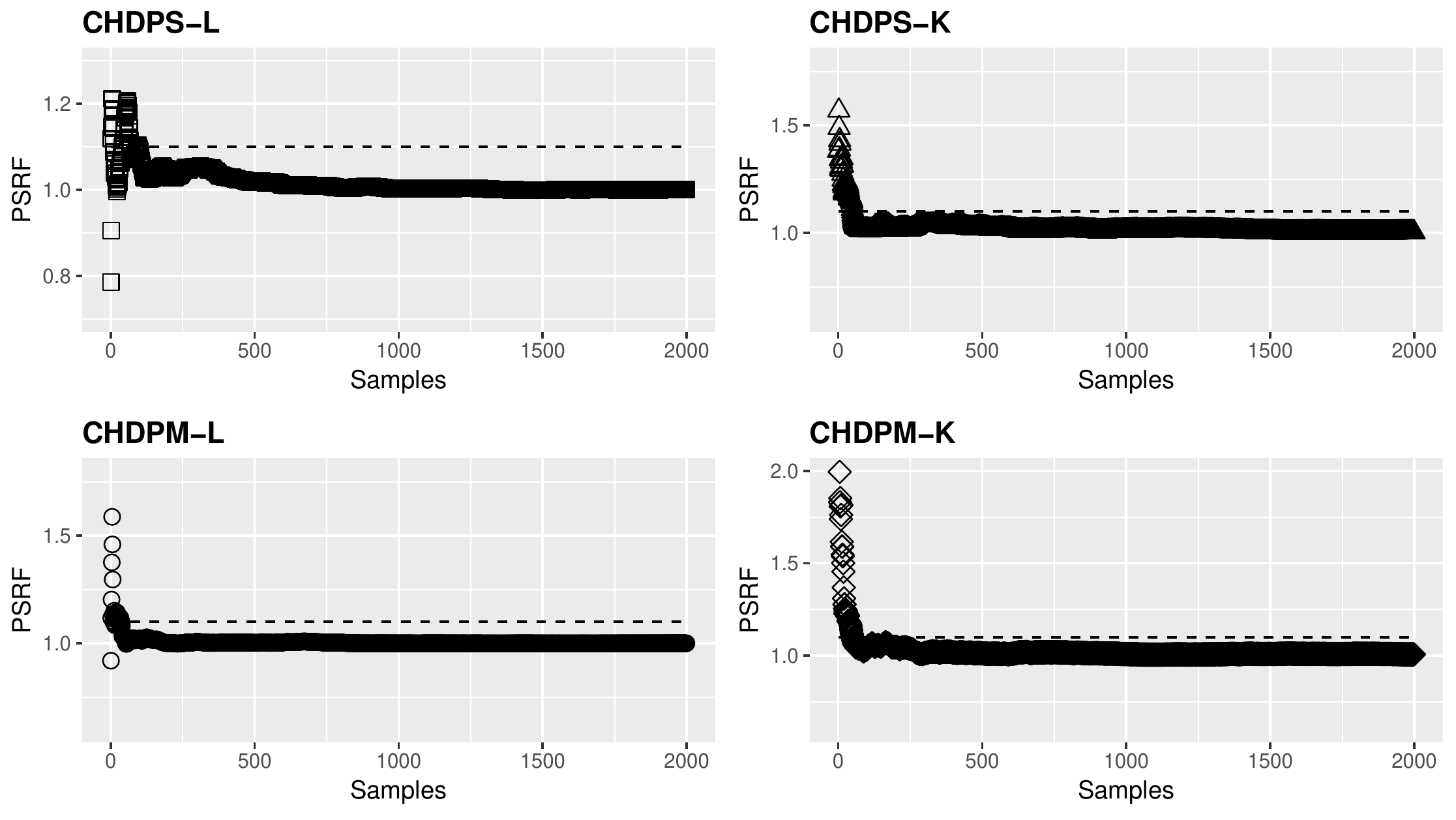}
\caption{Convergence comparison between CHDPS and CHDPM using PSRF on \emph{Likelihood}, \emph{K} and \emph{PSRF}. (There are 5 chains, and each chain contains 2,000 samples. Usually, it is acceptable that the sampler is convergent if PSRF is smaller than 1.2 or 1.1.)}
\label{chdp:psrf}
\end{figure}


We also evaluated the convergence on multiple (five) chains using the same synthetic data. The evaluation metric for multiple chains is the Potential Scale Reduction Factor (PSRF) \cite{gelman1992}, which is computed by $\sqrt{\frac{n-1}{n} + \frac{B}{nW}}$ where $B$ is the variance between the means of 5 chains, $W$ is the average of 5 within-chain variances, and $n=2000$ is the number of samples. Generally, the convergence is acceptable if PSRF is less than 1.2 or 1.1. Fig. \ref{chdp:psrf} shows the PSRF results of CHDPS and CHDPM on \emph{Loglikelihood} and \emph{K}. We also plotted a dashed line with PSRF=$1.1$ in each subfigure. From this figure, we can draw the following conclusions: 1) CHDPS and CHDPM both converged well because PSRF was smaller than 1.1 after about 500 samples; an 2) CHDPM converged more quickly than CHDPS because CHDPM-L  converged after about 200 samples but CHDPS-L used about 500 samples.

\subsection{Evaluation on parameter sensitivity}
\label{sec:parameters}

\begin{figure}[!t]
\centering
\includegraphics[scale=0.5]{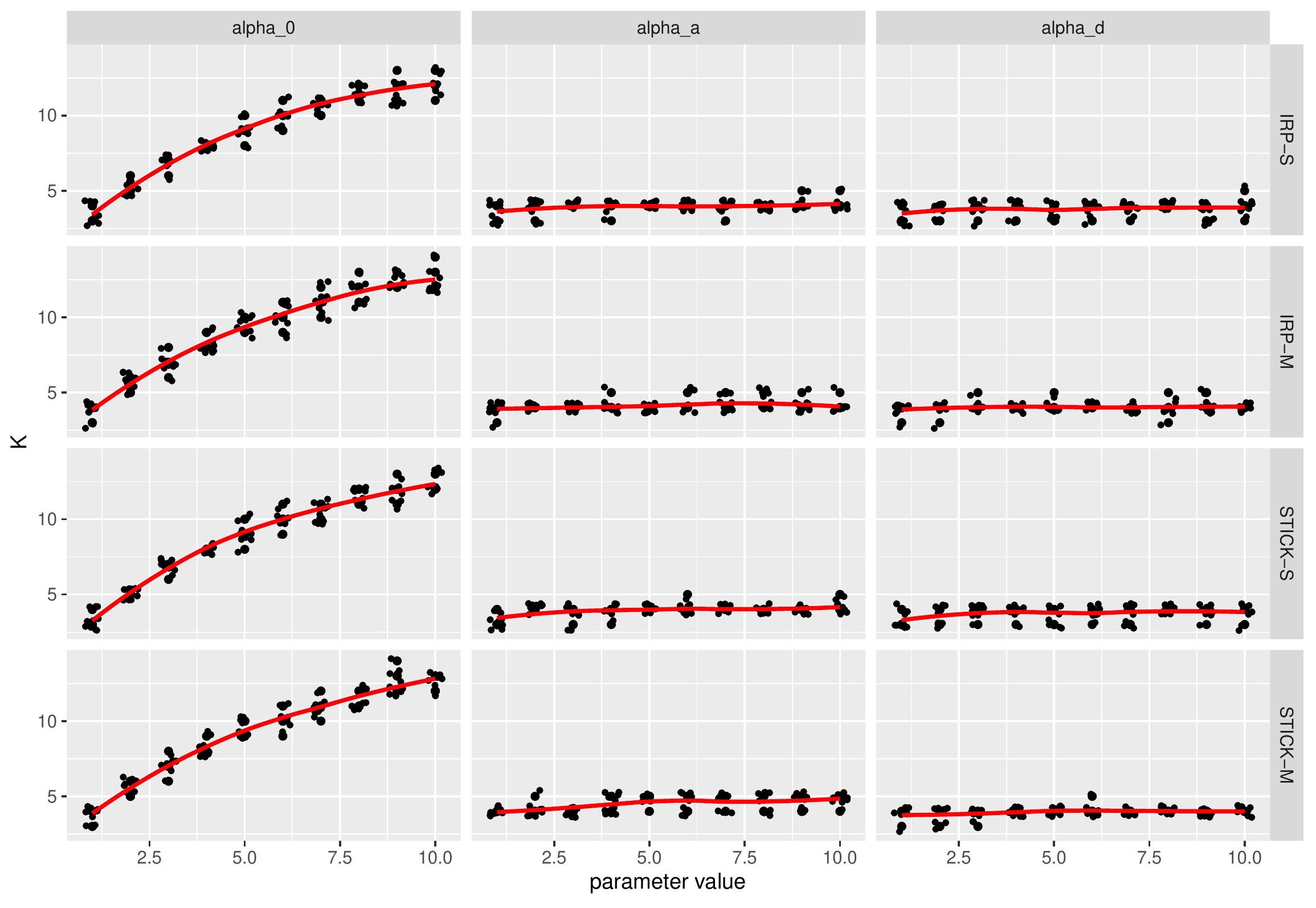}
\caption{The empirical evaluation on how the learned factor number $K$ from CHDP is changing with model parameters (i.e., $\alpha_0$, $\alpha_a$ and $\alpha_d$). IRP\_S denotes CHDPS under IRP representation; IRP\_M denotes CHDPM under IRP representation; STICK\_S denotes CHDPS under Stick-breaking representation; STICK\_M denotes CHDPM under Stick-breaking representation.}
\label{chdp:topicnum}
\end{figure}

The Bayesian nonparametric models (i.e., different stochastic processes or their designed combinations) actually provide a prior for the number of hidden factors. Given a dataset, we can infer a factor number for this dataset through Bayesian nonparametric models. The expected factor number from this prior is determined by the parameters of the designed nonparametric priors, so it is necessary to investigate the relationships between the model parameters with the inferred factor number. For the proposed CHDP, the expected factor number (including two representations: stick-breaking and IRP) is parameterized by $\alpha_0$, $\alpha_a$ and $\alpha_d$. In order to evaluate changing the factor numbers with three parameters, we first randomly generated a cooperative hierarchical structure with size $A=10$, $D=20$ and $V=50$. The links between nodes at three layers were also randomly set. The mixing density between $A$ and $D$ was set to 0.3 with a guarantee that each author is linked to at least one document and each document had at least one author, and the mixing density between $D$ and $V$ was set to 0.5 with a guarantee that each document is linked to at least one word and each word is linked to at least one document. We then ran both CHDPS and CHDPM (using IRP representation and Stick-breaking representation) on this generated cooperative hierarchical structure with different values of parameters and ignored the data likelihood, and recorded the final learned empirical factor number. Since we had three parameters $\alpha_0$, $\alpha_a$ and $\alpha_d$, we adjusted each one (taking a value from $\{1, 2, 3, 4, 5, 6, 7, 8, 9, 10\}$) with another two parameters fixed as $1.0$. In Figure \ref{chdp:topicnum}, there were $3 \times 3$ subfigures and each subfigure denoted a setting with one adjusted parameter, two fixed parameters, and a model under a specific representation. For example, the top-left corner subfigure had a setting: free $\alpha_0$, $\alpha_a=1$, $\alpha_d=1$, and IRP-S (i.e., CHDPS under IRP representation). For each candidate value of $\alpha_0$, we ran IRP-S 10 times and the learned hidden factor number at each time was represented as a (black) point in the subfigure. Furthermore, a trending (red) line of factor number changing with $\alpha_0$ was fitted and plotted. From this figure, we see that 1) CHDPS and CHDPM with two representations had similar trends of factor changes; 2) CHDP was more sensitive to $\alpha_0$ than $\alpha_a$ and $\alpha_d$, the reason being that $\alpha_0$ controls the factor number of the top level.

\begin{figure}[!t]
\centering
\includegraphics[scale=0.5]{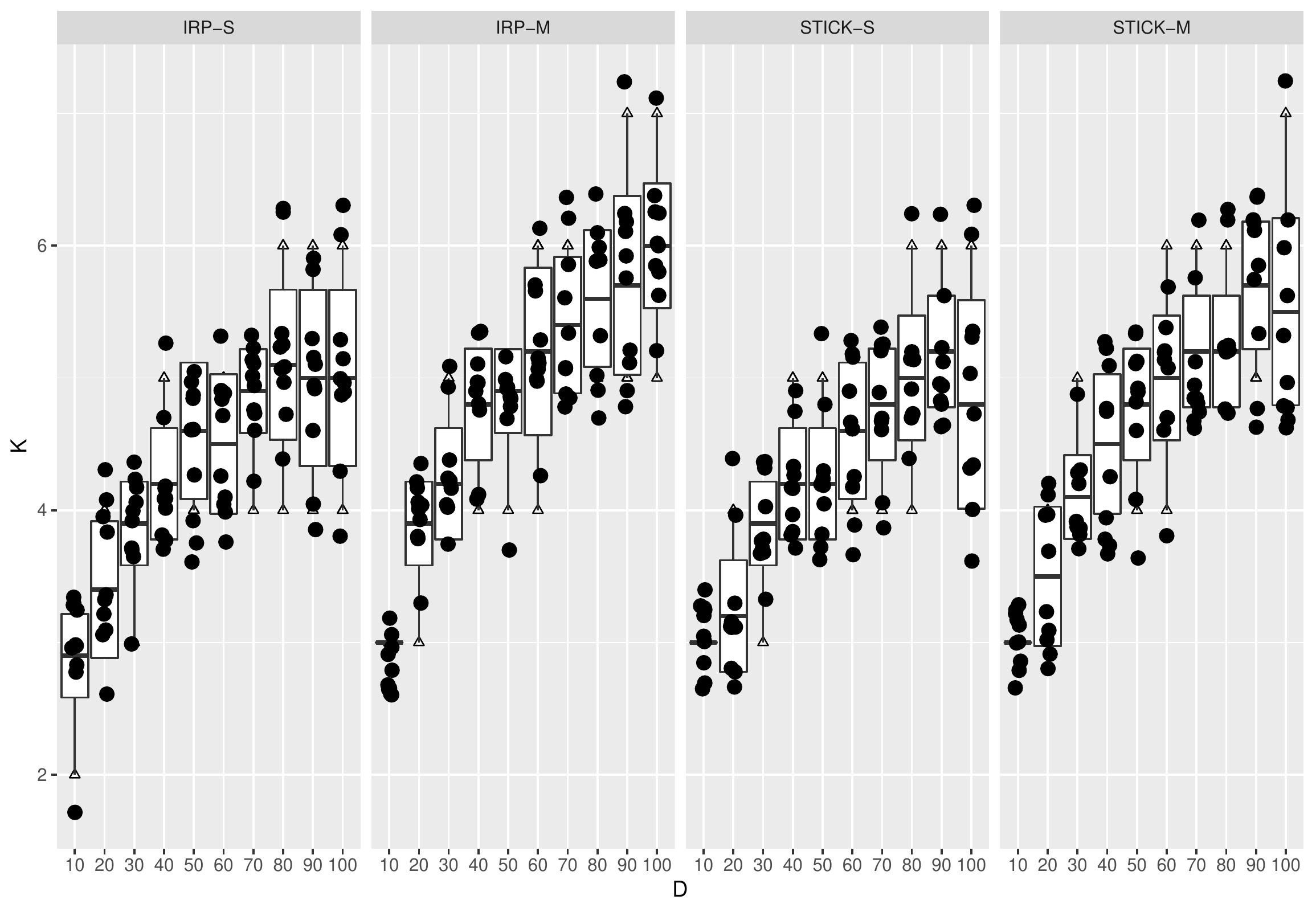}
\caption{The empirical evaluation on how the learned factor number $K$ from CHDP changes with data scale (i.e., the node number in hierarchical structure). IRP\_S denotes CHDPS under IRP representation; IRP\_M denotes CHDPM under IRP representation; STICK\_S denotes CHDPS under Stick-breaking representation; STICK\_M denotes CHDPM under Stick-breaking representation. }
\label{chdp:topicnumdata}
\end{figure}

After the relation between the hidden factor number and model parameters was evaluated, we were also interested in changes to the data scale (i.e., the node number in a hierarchical structure). A series of hierarchical structures were generated with a different number of nodes. For each hierarchical structure, we first fixed the number of nodes at the middle layer as $D$, and then the node number at the top layer was set as $A=\lfloor0.5*D\rfloor$ and the node number at the bottom layer was set as $V=D*2$. The mixing links between the top and middle layers were randomly generated with a fixed density of $0.3$ with a guarantee that each author is linked to at least one document and each document had at least one author, and the mixing links between the middle and bottom layers were also randomly generated with a fixed density of $0.5$. We ran CHDP under different representations on this series of hierarchical structures with the same parameters: $\alpha_0=1, \alpha_a =1, \alpha_d = 1$. All the results are shown in Fig. \ref{chdp:topicnumdata}. On each hierarchical structure, we ran the model 10 times, the learned factor number were represented as (black) points in Fig. \ref{chdp:topicnumdata}, and a box-plot was plotted to show the statistics of the factor numbers on this structure. From this figure, we see that: 1) CHDPS and CHDPM under two representations had similar but different trends for the hidden factor number changes; 2) CHDPM was relatively more sensitive to the data scale than CHDPS.

\subsection{Evaluation on cooperative structure modeling}
\label{sec:cooperative}

\begin{figure}[!t]
\centering
\includegraphics[scale=0.6]{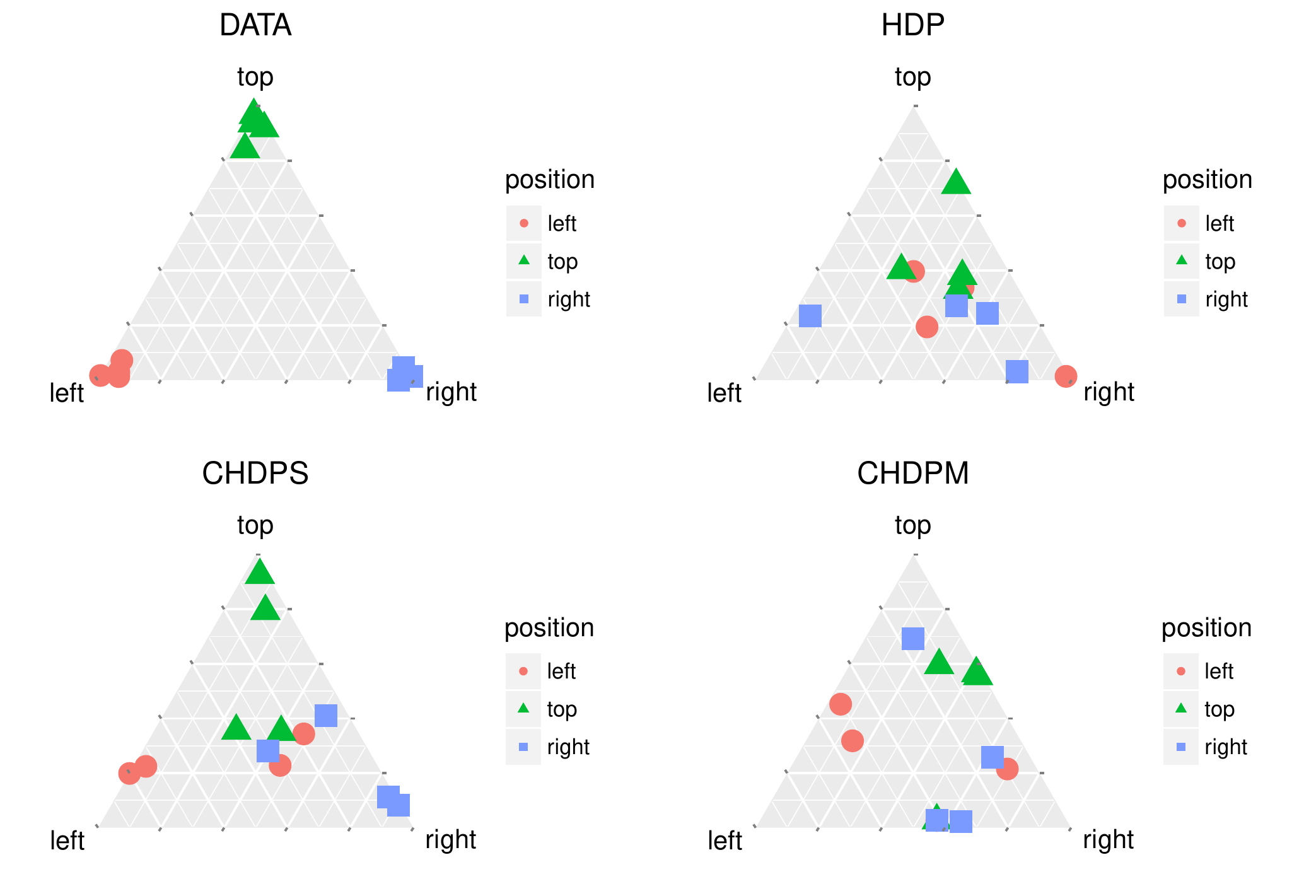}
\caption{Illustration of 12 authors' interests (points) on three vocabulary words, and same color and shape points denote authors in a group. Four subfigures denote: DATA (benchmark using Superposition), learned structure from HDP, learned structure from CHDPS, and learned structure from CHDPM. It appears that results from CHDPS are closer to the benchmark. Three quantitatively measured distances are: $<0.6256, 0.5631, 0.6399>$ (they are Euclidean distances and smaller value is better).}
\label{chdps:hdpcompone}
\end{figure}

The main contribution of this study is to extend HDP from a non-cooperative hierarchical structure (non-CHS) to a cooperative hierarchical structure (CHS). The main difference between non-CHS and CHS is the mixing relations in CHS. Next, we show the capability of CHDP on mixing structure modeling, comparing HDP using toy examples. Firstly, we generated 12 nodes (authors) at the top layer, 20 nodes (documents) at the middle layer, and 3 nodes (vocabulary words) at the bottom layer (here, we continue to use \emph{author-document-word} to explain CHS) as follows: 1) Authors were evenly divided into three groups and an author's interest ($v_a$ denoted the interest of author $a$) in three vocabulary words was generated by a group-specific Dirichlet distribution (parameterized by $<20, 1, 1>$, $<20, 1, 1>$, and $<20, 1, 1>$, respectively); 2) the mixing relations between authors and documents were randomly generated with fixed density of 0.3 with a guarantee that each author is linked to at least one document and each document has at least one author; 3) each document's interest in (three) vocabulary words was inherited from the cooperation (using Superposition) of its authors; 4) Finally, we generated 100 (maybe similar) words for each document using a multinomial distribution parameterized by its interest on three basic vocabulary words. Until now, we obtained a CHS, and we then ran CHDP (using IRP representation in Algorithm \ref{ag:irp}) on this CHS aiming to recover the authors' interests on three vocabulary words by $\pi_a * \theta$ (after normalization). At the same time, we degenerated CHS to a non-CHS by removing the redundant links between authors and documents to ensure each document had only one author, and then we ran HDP on this non-CHS to recover the authors' interests as well (all three models use the same parameters: $\alpha_0 = 1, \alpha_a = 1, \alpha_d = 1, \eta = 0.5$). If CHDP is able to recover the authors' interests better than HDP, this verifies that CHDP is able to model the mixing structure well because the only difference between CHS for CHDP and non-CHS for HDP is the mixing structure. Fig. \ref{chdps:hdpcompone} clearly demonstrates the results. There are four subfigures in Fig. \ref{chdps:hdpcompone}, and each subfigure has a 2-simplex which is a space for interests in three vocabulary words (each corner denotes a vocabulary word). The top-left subfigure shows the real author interests, where authors in same group are indicated by the same color and shape. The other three subfigures are results from HDP, CHDPS, and CHDPM, respectively. From this figure, we can see that 1) CHDPS could recover the hidden structure better than HDP (12 points in CHDPS were more closer to their real positions in DATA than them in HDP) because it had considered the mixing structure; and 2) CHDPS was better than CHDPM because the data was generated using \emph{Superposition} rather than \emph{Maximization}. Note that we also measured the (Euclidean\footnote{We also tried other distances, e.g., cosine and correlation, finding that they have same trend as Euclidean.}) distances between the real (DATA in Fig. \ref{chdps:hdpcompone}) and learned positions of the authors quantitatively except for the visualization in Fig. \ref{chdps:hdpcompone}: $0.6256$ for HDP, $0.5631$ for CHDPS, and $0.6399$ for CHDPM.

\begin{figure}[!t]
\centering
\includegraphics[scale=0.6]{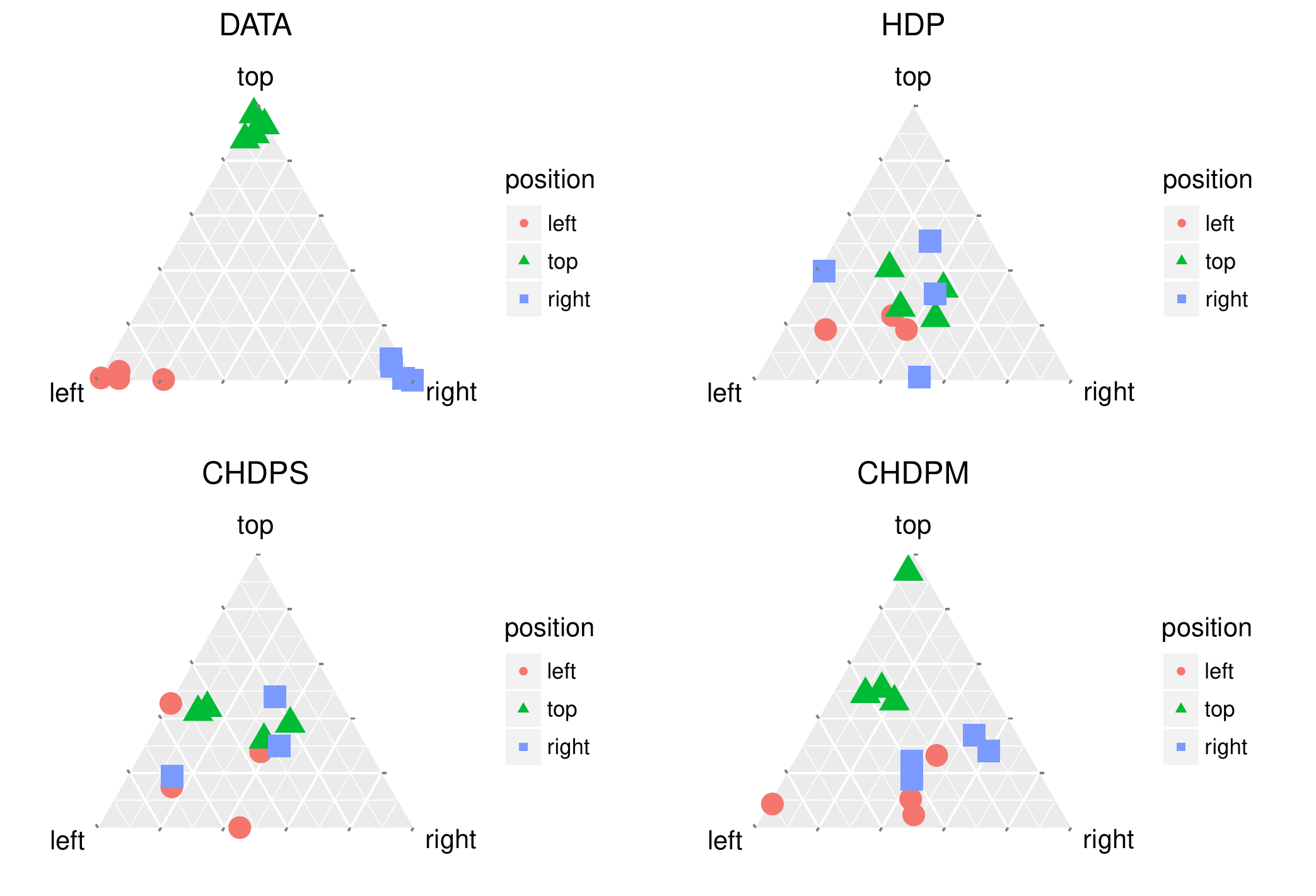}
\caption{Illustration of 12 authors' interests (points) in three vocabulary words, where the same color and shape points denote authors in a group. The four subfigures denote: DATA (benchmark using Maximization), the learned structure from HDP, the learned structure from CHDPS, and the learned structure from CHDPM. It can be seen that the results from CHDPM are closer to the benchmark. Three quantitatively measured distances are: $<0.6940,0.6440,0.4915>$ (these are Euclidean distances and a smaller value is better).}
\label{chdpm:hdpcompone}
\end{figure}

The reason why CHDPS was better than CHDPM in the above example is because the data was generated using \emph{Superposition}. To prove this argument, we generated another toy dataset using the same procedure with only one difference in step 3: each document's interest in (three) vocabulary words was inherited from the cooperation (using \emph{Maximization} rather than \emph{Superposition}) of its authors. We then performed this evaluation again using the same settings, and the results were shown in Fig. \ref{chdpm:hdpcompone}. From this figure, we can see that: 1) CHDPM recovered the hidden structure better than HDP; 2) CHDPM was also better than CHDPS on this toy example. The Euclidean distances were: $ 0.6940$ for HDP, $0.6440$ for CHDPS, and $0.4915$ for CHDPM. One interesting observation was that the performance of CHDPS was a little worse than HDP in the toy example using \emph{Maximization} and the performance of CHDPM was also a little worse than HDP in the toy example using \emph{Superposition}. This observation tells us that CHDPS and CHDPM are not interchangeable and both are necessary because we had no knowledge about how the real-world data were generated. Furthermore, it demonstrated that choosing the appropriate model is a determinant for learning CHS and the performance of a wrong model may be even worse than ignoring a cooperative structure.

\begin{figure}[!t]
\centering
\includegraphics[scale=0.7]{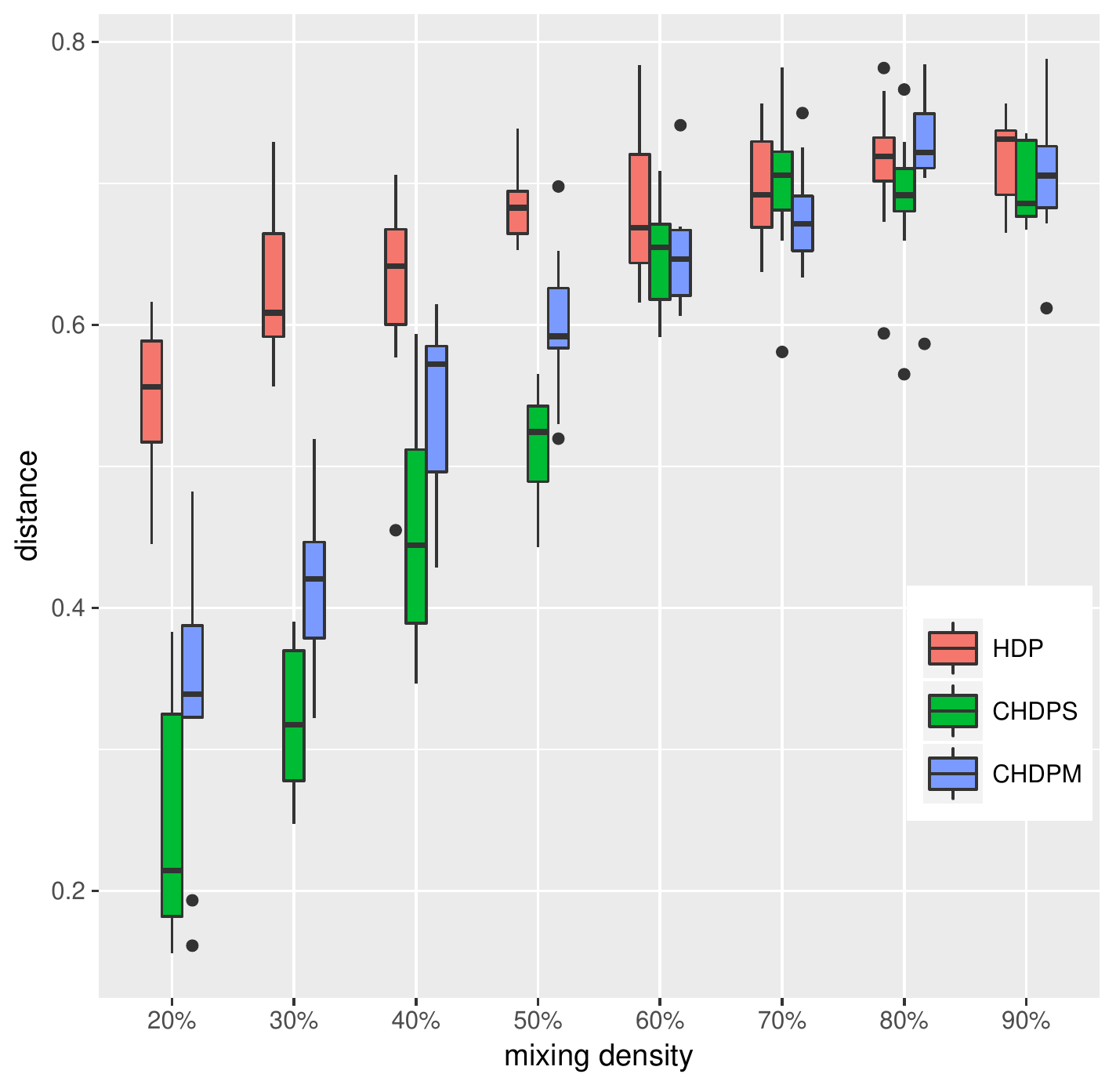}
\caption{The evaluation on the capability of cooperative structure modeling with different mixing densities. For each density, the synthetic data (using Superposition) is simulated 10 times and three models also run 10 times, so three box-plots at each density summarize the results of the three models.}
\label{chdps:hdpcompall}
\end{figure}

\begin{figure}[!t]
\centering
\includegraphics[scale=0.7]{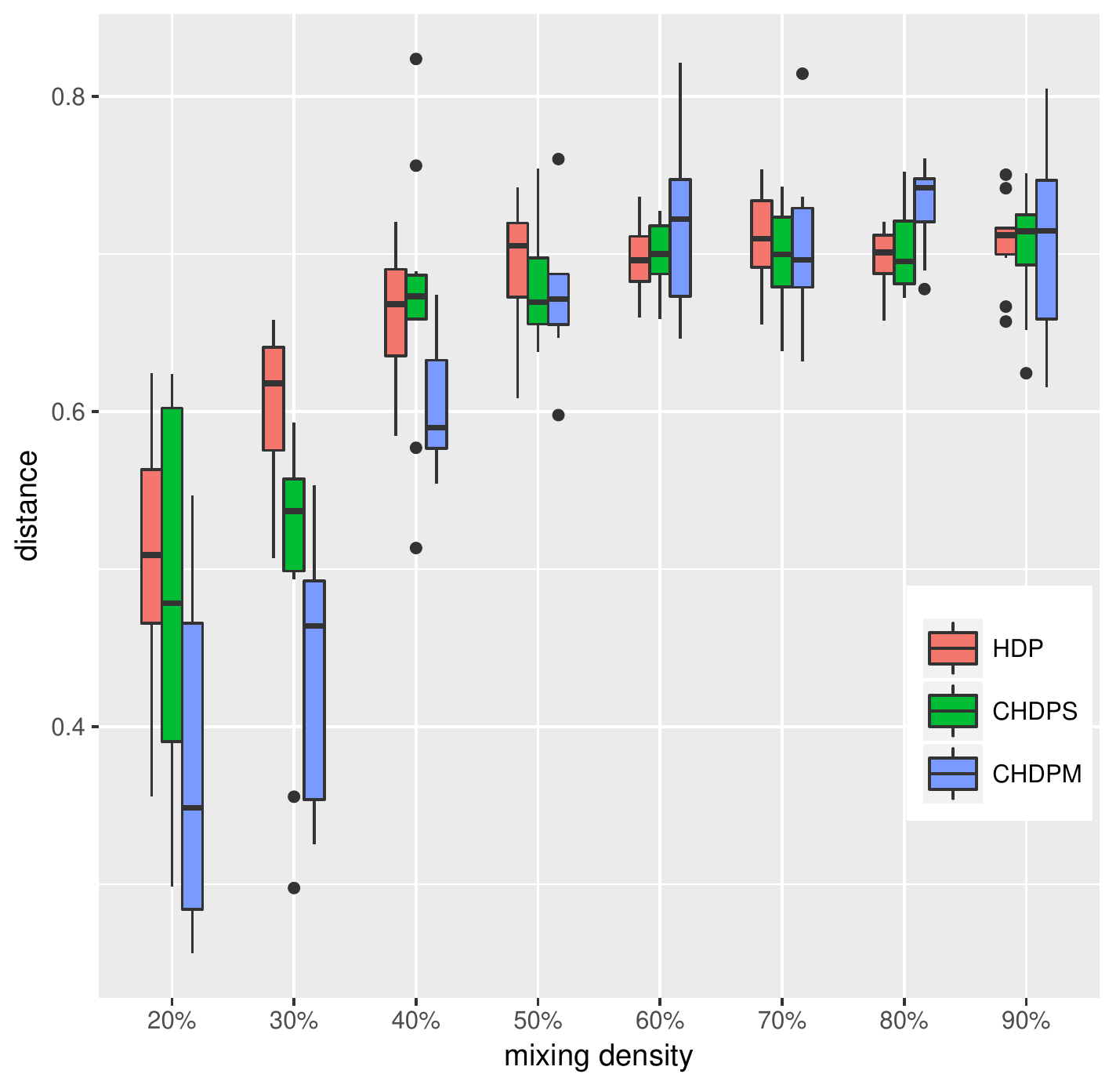}
\caption{The evaluation on the capability of cooperative structure modeling with different mixing densities. For each density, the synthetic data (using Maximization) is simulated 10 times and three models also run 10 times, so three box-plots at each density summarize the results of the three models. }
\label{chdpm:hdpcompall}
\end{figure}

The capability of CHDP on cooperative structure modeling has been evaluated and analyzed using the aforementioned two examples with fixed mixing density (0.3) between authors and documents. It is also interesting to evaluate its performances with different mixing densities. We used the above data generation procedure, but the density was adjusted with values of \{0.2, 0.3, 0.4, 0.5, 0.6, 0.7, 0.8, 0.9\}. For each density, we repeated the following process 10 times: 1) simulated a data; and 2) ran three models (i.e., HDP, CHDPS, and CHDPM) using same model parameters as above. As before, we had also considered both \emph{Superposition} and \emph{Maximization} respectively. Figs. \ref{chdps:hdpcompall} and \ref{chdpm:hdpcompall} show the results on the data using \emph{Superposition} and \emph{Maximization}, where the x-axis denoted mixing density and the y-axis denoted the distance between learned authors' interests from the three models to the real authors' interests (similar to the aforementioned toy examples). At each density, there were three box-plots corresponding to the three models (pink for HDP, green for CHDPS, and blue for CHDPM), and each box-plot was used to summarize 10 points/results from a specific model in both figures. From Fig. \ref{chdps:hdpcompall} and \ref{chdpm:hdpcompall}, we see that: 1) CHDPS and CHDPM generally performed better than HDP; 2) CHDPS was better than CHDPM on hidden structure learning using \emph{Superposition}, and CHDPM was better than CHDPS on the hidden structure learning using \emph{Maximization}; 3) After increasing the density, the performance of all models decreased, which was due to an increase in the complexity of the mixing relations; 4) more interestingly, the performance of the three models was indistinguishable when the density went beyond 0.6 or 0.5. The underlying reason for this is the identification problem\footnote{More detail on this problem can be found in http://www2.gsu.edu/\~mkteer/identifi.html} where it is impossible to distinguish the respective contributions of authors in the extreme situation that density is 1.0 (all authors write all documents together). The mixing structure between authors and documents will determine if the authors' interests can be identified or not. We believe this problem will appear if the rank of the mixing matrix is significantly smaller than the number of authors, so it is necessary to check this factor before using the proposed models or HDP.

\subsection{Evaluation on real-world tasks}
\label{sec:realworld}

Following the previous evaluations on the model properties of CHDP using synthetic data, this subsection evaluates the capability of CHDP to resolve real-world tasks using real-world datasets. The two selected document-based real-world tasks are: \emph{Author-topic modeling} and \emph{Multi-label classification}. The reason these are selected is that both tasks involve cooperative hierarchical structures. The variational inference in Algorithm \ref{ag:vi} is adopted for both tasks. Next, we introduce each task in more detail, including the dataset, aim, comparative models, setup, evaluation metric, and the result analysis.

\subsubsection{Author-topic modeling task}

\begin{figure}[!t]
\centering
\includegraphics[scale=0.65]{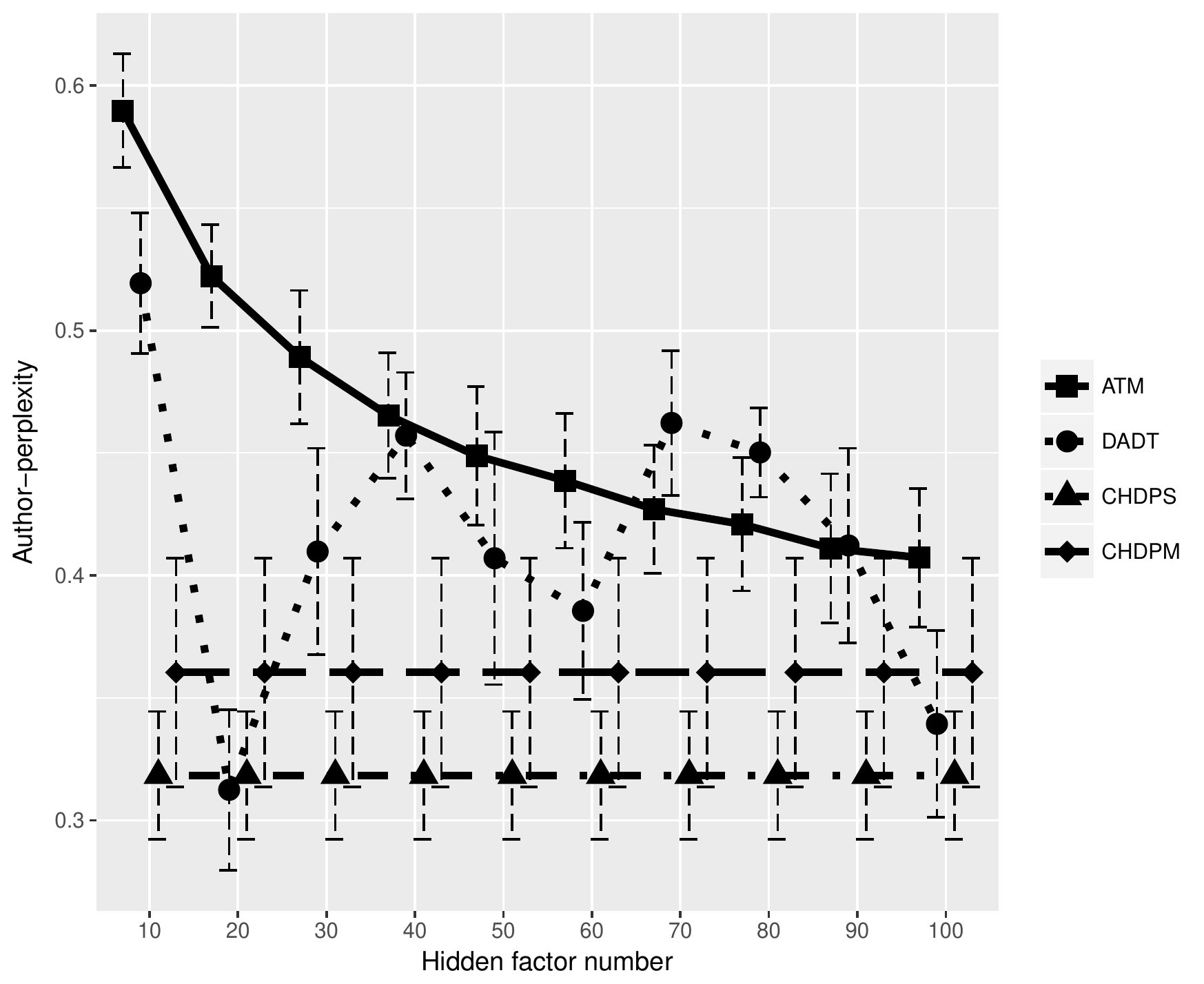}
\caption{The evaluation on the \emph{author-topic modeling} task. The two selected comparative Bayesian models (with fixed dimensions) are ATM and DADT, and the evaluation metric is $Author$-$perplexity$. The x-axis denotes the candidates of hidden topic numbers, and the results from the four models on each candidate using 5-fold cross-validation are plotted including mean and standard deviation. }
\label{exp:authperp}
\end{figure}

The \textbf{DATASET} for this task is \textit{NIPS papers}\footnote{http://www.datalab.uci.edu/author-topic/NIPs.htm}. This dataset contains papers from the NIPS conferences between 1987 and 1999, comprising 1,740 papers with 2,037 authors, a total of 2,301,375 word tokens, and a vocabulary size of 13,649 unique words. Note that this dataset is actually a \textbf{COOPERATIVE HIERARCHICAL STRUCTURE}: \emph{author-paper-word} (Each paper could have more than one author), so the proposed CHDP could be adopted to model this dataset. The \textbf{AIM} of this task is to discover the hidden topics/factors from this structure and, simultaneously, the authors' interest in these topics. This task could be further applied to real-world applications, such as 1) detecting the most and least surprising papers for an author, 2) an author/topic-based browser; and so on. The selected \textbf{COMPARATIVE MODELS} for this task are the \textit{Author Topic Model (ATM)}\footnote{Implementation is from: http://www.datalab.uci.edu/author-topic/} \cite{atm1} and the \textit{Disjoint Author-Document Topic model (DADT)}\citep{Seroussi:2014:AAT}, which are based on fixed dimensional probability distributions. Note that the topic number needs to be fixed when using ATM and DADT, but CHDP does not suffer from this problem. The \textbf{SETUP} for this task was as follows: 5-fold cross validation was applied so the entire dataset was divided into 5 parts, with one being used as the test data each time. Furthermore, the rank of the mixing matrix between labels and free texts is around 1107 for each fold, which is close to the rank maximum. After learning the proposed models on the training dataset, we predicted the authors of a given test paper. CHDP was implemented using the stick-breaking representation with both Superposition and Maximization in Section \ref{sec:stick} and Algorithm \ref{ag:vi}. CHDP used the following truncation levels: $T=50, O=100, K=500$; and parameters: $\alpha_0 = 1, \alpha_a = 1, \alpha_d = 1, \eta = 0.5$. The \textbf{EVALUATION METRIC} used for the qualitative comparisons is \emph{Author-perplexity}: $Ap = \exp\left( -\frac{1}{|\mathfrak{D}^t|}\sum_{d \in \mathfrak{D}^t} \frac{1}{A_d}\sum_{a\in a_d} \sum_k p(a|\theta_k) p(\theta_k|w_d) \right)$, where $\mathfrak{D}^t$ is the test papers, $\theta_k$ is the learned $k$-th topic, $a_d$ is the authors of paper $d$, and $A_d$ is the author number of paper $d$. $Ap$ is the exponential of the probability of observing authors $a_d$ of a given document $d$. The smaller the value of $Ap$, the better the performance. For CHDP, $p(a|\theta_k)$ can be evaluated by $\pi_{a,k} = \sum_{o: z_{a,o}=k} \pi_{a,o}$, and $p(\theta_k|w_d)$ can be evaluated by the cosine distance between $\theta_k$ and $w_d$. For CHDPM, the evaluation is a little different: $Ap = \exp\left( -\frac{1}{|\mathfrak{D}^t|}\sum_{d \in \mathfrak{D}^t} \sum_k p(\widetilde{a}|\theta_k) p(\theta_k|w_d) \right)$, where $\widetilde{a}$ is the \emph{Maximization} of all author interests of paper $d$. The \textbf{RESULTS} are shown in Figure \ref{exp:authperp}. Since ATM and DADT need the number of topics to be fixed in advance, the 10 candidates $K \in \{i: i = j \times 10, j = [1, 10]\}$ (indicated by the x-axis) were evaluated and plotted in Figure \ref{exp:authperp}. Since CHDPS and CHDPM do not have this limitation, there were two lines in the figure to represent their results. We also plotted the standard deviations from the cross-validation. From Figure \ref{exp:authperp}, we can see that 1) the performances of ATM and DADT were affected by choosing the hidden topic number; 2) CHDPS and CHDPM achieved generally better performances than ATM and DADT. Note that CHDP achieved this better performance without the additional restriction that the topic number needs to be prefixed; 3) CHDPS was slightly better than CHDPM, and it was interesting that CHDPM had a comparative performance to CHDPS on this task; and 4) the standard deviations from CHDPM were the largest of all the models, which may be because \emph{Maximization} in CHDPM is not a strict restriction on all authors' interests compared to \emph{Superposition} and this loose restriction leaded more variance. So, we can draw the conclusion that CHDP is effective on this task.

\subsubsection{Multi-label classification task}

\begin{figure}[!t]
\centering
\includegraphics[scale=0.55]{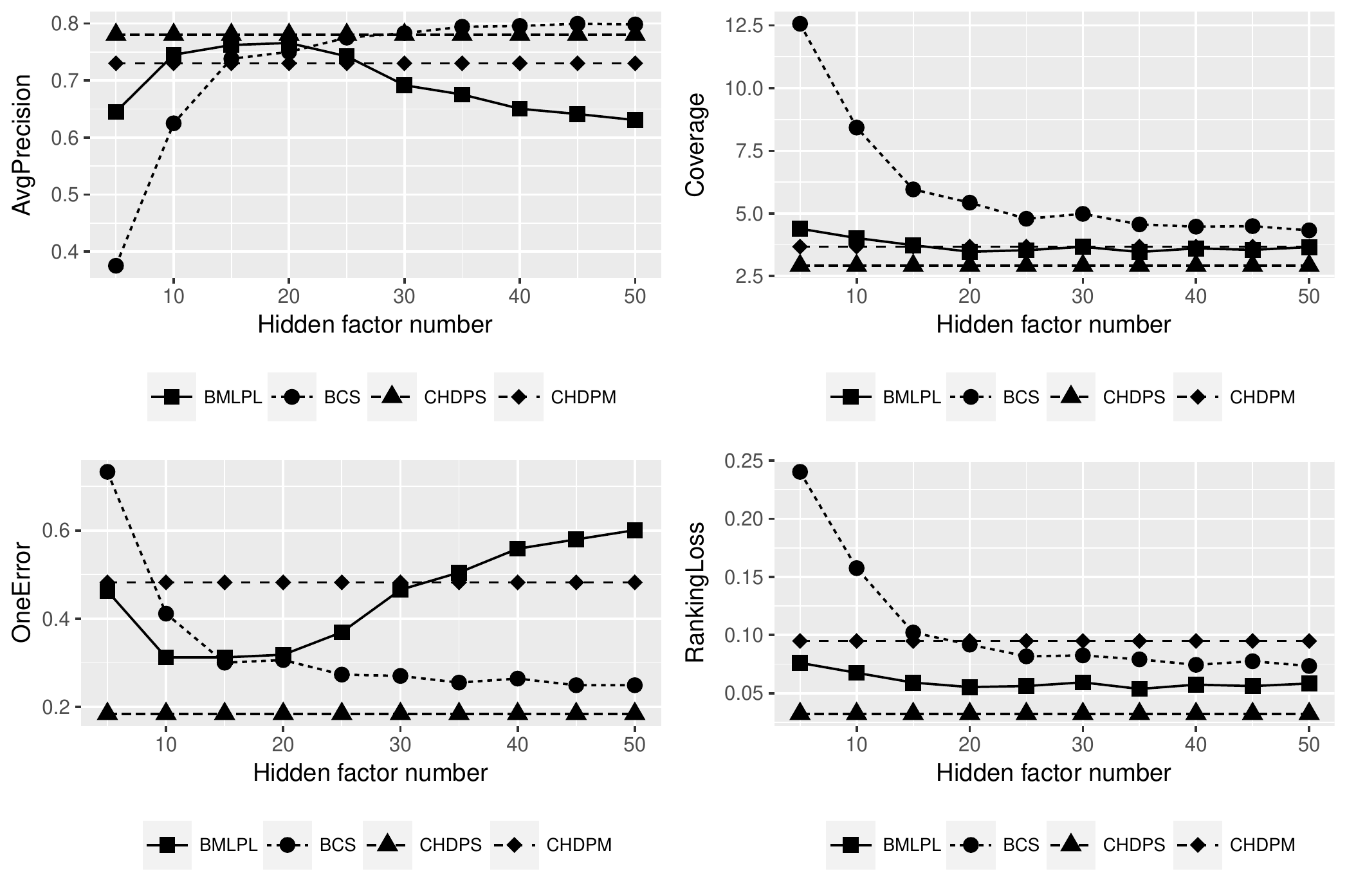}
\caption{The evaluation on the \emph{multi-label classification} task. The two selected comparative Bayesian models (with fixed dimensions) are BCS and BMLPL. The four subfigures denote the four evaluation metrics. The x-axis denotes the candidates of the hidden factor numbers, and the results from the four models on each candidate are plotted. }
\label{exp:multilabel}
\end{figure}

The \textbf{DATASET} for this task is \textit{Clinical free text}\footnote{http\://mulan.sourceforge.net/datasets-mlc.html}. This dataset comprises radiology reports annotated by experts. There are 45 labels (ICD-9-CM codes) and 645 (training) / 333 (testing) texts with 1,449 features. More detailed description can be found in \cite{pestian2007shared}. Note that this dataset is also a \textbf{COOPERATIVE HIERARCHICAL STRUCTURE}: \emph{label-text-feature} (Each clinical text may have more than one label), so the proposed CHDP can also be adopted to model this dataset. Since each text is associated with a number of (0/1 valued) features (similar to mapping between documents and words), the multinomial distribution is still used as the likelihood of CHDP for this dataset. Furthermore, the mixing matrix between labels and free texts has a full rank. The \textbf{AIM} of this task is to automatically assign labels to the test clinical texts. Automatic and accurate label assignment for text can save an enormous amount of time and cost compared with manual labor. The selected \textbf{COMPARATIVE MODELS} for this task are \textit{Bayesian Compressed Sensing (BCS)}\footnote{Implementation is from: https://github.com/yalesong/BGCS} \citep{KapoorVJ12} and \textit{Bayesian Multi-label Learning via Positive Labels (BMLPL)}\footnote{Implementation is from: http://people.ee.duke.edu/~lcarin/Papers.html}  \citep{NIPS20155770} (two Bayesian models with fixed dimensions). Note that the factor number needs to be fixed when using BCS and BMLPL, but CHDP does not suffer from this problem. The \textbf{SETUP} for this task was as follows: we trained CHDP using training data and learned the hidden factor embedding for labels and features, and then used this factor embedding to predict the labels for the test dataset. CHDPS and CHDPM used the following truncation levels: $T=50, O=100, K=200$; and parameters: $\alpha_0 = 1, \alpha_a = 1, \alpha_d = 1, \eta = 0.1$. The \textbf{EVALUATION METRICS} are: \emph{OneError}, \emph{Coverage}, \emph{RankingLoss}, and \emph{AvgPrecision}, which are commonly used for a performance comparison of multi-label learning and their detailed definitions can be found in \cite{Madjarov20123084}. For \emph{AvgPrecision}, the larger the value, the better the performance; for \emph{OneError}, \emph{Coverage} and \emph{RankingLoss}, the smaller the value, the better the performance. These metrics are all ranking-based. This means that they can rank all the labels in different multi-label classification models for every data according to the possibility of the data with each label. For CHDP, it can also rank the labels of the test data according to their hidden factor embedding by $Rank(l, x_i) = <\pi_{l}, \pi_{x_i}>$, where $x_i$ is $i$-th test data, $l$ denotes a label, $\pi_{l}$ is a $K$ dimensional vector that denotes the factor embedding of label $l$ and $\pi_{l, k}$ can be evaluated by $\pi_{l,k} = \sum_{o: z_{l,o}=k} \pi_{l,o}$, $\pi_{x_i}$ is also a $K$ dimensional vector that denotes the factor embedding of data $x_i$ and $\pi_{x_i, k}$ can be evaluated by the cosine distance between $\theta_k$ and $x_i$. Finally, we can rank the labels for each data according to $Rank(l, x_i)$.
The \textbf{RESULTS} are shown in Figure \ref{exp:multilabel}, where four subfigures denote the four evaluation metrics and there are four lines plotted for the four models in each subfigure. Since BCS and BMLPL need the number of topics to be fixed in advance, the 10 candidates $K \in \{i: i = j \times 5, j = [1, 10]\}$ (indicated by the x-axis) were evaluated and plotted in each subfigure. The results from CHDPS and CHDPM were again represented as two straight-lines in each subfigure. From Figure \ref{exp:multilabel}, we observed that 1) the performances of BCS and BMLPL fluctuated with the hidden factor numbers; 2) CHDPS achieved the best performance on \emph{OneError} and \emph{RankingLoss}, and achieved a comparative performances on \emph{AvgPrecision} with BCS and \emph{Coverage} with BMLPL; 3) CHDPM performed badly on \emph{OneError} and \emph{RankingLoss}, but achieved comparative performances on \emph{AvgPrecision} and \emph{Coverage}. So, we can draw the conclusion that CHDPS is effective on this task and CHDPS is better than CHDPM on this task.

\section{Conclusions and further studies}

Hierarchical structure is a commonly observed and adopted data structure, so its modeling could benefit numerous application areas, such as author-topic modeling and multi-label learning. We have presented a Bayesian nonparametric model, i.e., cooperative hierarchical Dirichlet processes (CHDP), for more general hierarchical structure: cooperative hierarchical structures. CHDP is based on two random measure operations which have been specifically designed to model the cooperative hierarchical structure (CHS): \emph{Inheritance} for the layering structure in CHS, \emph{Cooperation: Superposition} and \emph{Cooperation: Maximization} for the mixing structure in CHS. Similar to the renowned DP and HDP, two constructive representations, i.e., the international restaurant process and stick-breaking, have been designed for CHDP to facilitate the model inference. In order to resolve the issue brought about by \emph{Inheritance} and \emph{Cooperation} in CHDP, two inference algorithms have been carefully developed for both representations. Experiments on synthetic and real-world datasets showed its ability to model cooperative hierarchical structures and demonstrated its practical application scenarios.

In the future, we plan to design a more efficient and accurate inference algorithm for CHDPM based on evolutionary computing considering its complicated non-smooth optimization objective function. Moreover, it would be also interesting to apply the idea of existing various extensions for HDP on CHDP accounting for more general situations. Other interesting work is to extend the current model to hierarchical network structures that include node network structures within each single layer of CHS.

\section*{Acknowledgements}

Research work reported in this paper was partly supported by the Australian Research Council (ARC) under Discovery Grant DP140101366.

\section*{References}

\bibliography{CHDP}

\section*{Appendix 1: Marginal sampler}

\begin{algorithm}[!t]
\caption{Marginal Sampler for IRP}
\label{ag:irp}
initialization\;
\Do{convergent}{
    \For{$d=1; d \le D$}
    {
        \For{$n=1; n \le N_d$}
        {
             Update $\theta_{d,n}$ by Eq. (\ref{irp:thetadn});\\
        }
        \For{$t=1; t \le T_d$}
        {
            // \emph{Superposition}\\
            Update $\theta_{d,t}$ by Eq. (\ref{irps:thetadt});\\
            // \emph{Maximization}\\
            Update $\theta_{d,t}$ by Eq. (\ref{irpm:thetadt});
        }
    }
    \For{$a=1; a \le A$}
    {
        \For{$o=1; o \le O_a$}
        {
            Update $\theta_{a,o}$ by Eq. (\ref{irp:thetaao});
        }
    }
    \For{$k=1; a \le K$}{
        Update $\theta_k$ by Eq. (\ref{irp:thetak});
    }
}
return $K$, $\{\theta_k\}_{k=1}^{K}$, $\{\{\theta_{d,t}\}_{t=1}^{T_d}\}_{d=1}^D$\}, $\{\{\theta_{a,o}\}_{o=1}^{O_a}\}_{a=1}^A$ \;
\end{algorithm}

\textbf{Sampling $\theta_{d,n}$.} To assign a table $t$ to each customer $n$ in restaurant $d$, the prior is as the one in Eq. (\ref{irp:thetadn}) and the likelihood part is
\begin{equation}\label{irp:lhthetadn}
\begin{aligned}
L(\eta_{d,t} ) \propto
\begin{cases}
    F(v_{d,n}| \theta_{d,t})  &\text{if $t$ is occupied} \\
    LK(v_{d,n})  &\text{if $t$ is new}
\end{cases}
\end{aligned}
\end{equation}

\textbf{Sampling $\theta_{a,o}$.} To assign a dish $k$ to each menu option $o$ of chef $a$, the prior is as the one in Eq. (\ref{irp:thetaao})
and the likelihood part is,
\begin{equation}\label{irp:lhthetaao}
\begin{aligned}
L(\eta_{a,o} ) \propto
\begin{cases}
    F(v_{a,o}| \theta_k)  &\text{if $k$ is occupied} \\
    LK(v_{a,o})  &\text{if $k$ is new}
\end{cases}
\end{aligned}
\end{equation}
where $v_{a,o}$ denotes all the customers served by the $o$-th menu option of chef $a$,
\begin{equation*}
\begin{aligned}
LK(x_{a,o})
&= \int_{\theta} F(v_{a,o}| \theta) H(\theta) \mathrm{d} \theta.
\end{aligned}
\end{equation*}

\textbf{Sampling} $\mathbf{\theta_k}$. $\theta_k$ denotes a global factor/topic and its posterior distribution is
\begin{equation}\label{irp:thetak}
\begin{aligned}
p(\theta_k | \cdots) &\propto Dir(\theta_k; \gamma) \cdot F (v_{k} | \theta_k)
\end{aligned}
\end{equation}
where $v_{k}$ is total number of customers assigned to $k$.

We can also introduce an auxiliary variable $\hat{z}_{d,n}$ to make it inferrable: $\hat{z}_{d,n}$ denotes the selected chef of customer $n$ in restaurant $d$. If we know which chef this customer selects, we can simply assign a dish to him by marginalizing the probability measure of the selected chef. Here, we define the distribution of the auxiliary variable $\hat{z}_{d,n}$ as
\begin{equation}\label{irp:zdn}
\begin{aligned}
p(\hat{z}_{d,n} = a | \cdots) = \frac{ \sum_t N^a_{d,t} + \alpha_d}{\sum_t N_{d,t} + \alpha_d}
\end{aligned}
\end{equation}
where $N^a_{d,t}$ denotes the number of customers on table $t$ served by chef $a$ in restaurant $d$. With the selected chef, we can sample $\theta_{d,n}$ by
\begin{equation}\label{irp:thetadnnew}
\begin{aligned}
\theta_{d,n} | \hat{z}_{d,n} = a, G_a, \cdots  \sim \sum_{t=1}^{T_d^a}
\frac{N_{d,t}}{\sum_t N^a_{d,t} + \alpha_d} \delta_{\theta_{d,t}} +
\frac{ \alpha_d}{\sum_t N^a_{d,t} +  \alpha_d} G_a
\end{aligned}
\end{equation}
where $T^a_d$ is the table number in restaurant $d$ served by chef $a$ and $t \in a$ denotes table $t$ served by chef $a$. If a new dish is needed, we need to sample from $G_a$.

\begin{prf}
The marginal distribution of $\theta_{d,n}$ with $\hat{z}_{d,n}$ marginalized out is:
\begin{align}
p(\theta_{d,n})
&= \sum_{\hat{z}_{d,n}} p(\theta_{d,n}, \hat{z}_{d,n}) p(\hat{z}_{d,n}) \notag\\
&= \sum_a p(\theta_{d,n}| \hat{z}_{d,n} = a) p(\hat{z}_{d,n} = a)  \notag\\
&= \sum_a
\left (\sum_{t=1}^{T_d^a}
\frac{N_{d,t}}{\sum_t N^a_{d,t} + \alpha_d} \delta_{\theta_{d,t}} +
\frac{\alpha_d}{\sum_t N^a_{d,t} +  \alpha_d} G_a \right )
 \frac{\sum_t N^a_{d,t} + \alpha_d}{\sum_t N_{d,t}  + \alpha_d}  \notag\\
&= \sum_a
\left (\sum_{t=1}^{T_d^a}
\frac{N_{d,t}}{\sum_t N_{d,t}  + \alpha_d} \delta_{\theta_{d,t}} +
\frac{ \alpha_d}{\sum_t N_{d,t}  + \alpha_d} G_a \right )
\notag\\
&=
\sum_{t=1}^{T_d}
\frac{N_{d,t}}{\sum_t N_{d,t}  + \alpha_d} \delta_{\theta_{d,t}} +
\frac{\alpha_d}{\sum_t N_{d,t}  + \alpha_d} \left ( G_{a_1} \oplus  G_{a_2} \oplus  \cdots \right )\notag\\
&=
\sum_{t=1}^{T_d}
\frac{N_{d,t}}{\sum_t N_{d,t}  + \alpha_d} \delta_{\theta_{d,t}} +
\frac{\alpha_d}{\sum_t N_{d,t}  + \alpha_d} G^d_a \notag
\end{align}

The result is the same as in Eq. (\ref{irp:thetadn}). So we can conclude that introducing an auxiliary variable will not impact on the posterior distribution of the $\theta_{d,n}$.
\end{prf}

\textbf{Sampling $\theta^a_{d,t}$}. To assign an menu option $o$ to each table served by chef $a$ in restaurant $d$, the prior for $\theta^a_{d,t}$ is as the one in Eq. (\ref{irp:thetadt})
and the likelihood part is,
\begin{equation}\label{irp:lhthetaadt}
\begin{aligned}
L(\theta^a_{d,t}) \propto
\begin{cases}
    F(v_{d,t}| \theta_{a, o})  &\text{if $o$ is occupied} \\
    \sum_{k=1}^{K}
\frac{O_{k}}{\sum_{k}{O_{k}} + \alpha_0} F(v_{d,t}| \theta_k) +
\frac{\alpha_0}{\sum_{k}{O_{k}} + \alpha_0} LK(v_{d,t})  &\text{if $o$ is new}
\end{cases}
\end{aligned}
\end{equation}
where $v_{d,t}$ denotes all the customers sitting on table $t$ in restaurant $d$, and $O_k$ is the number of menu options with dish $k$,
\begin{equation*}
\begin{aligned}
LK(v_{d,t})
&= \int_{\theta} F(v_{d,t}| \theta) H(\theta) \mathrm{d} \theta.
\end{aligned}
\end{equation*}

\section*{Appendix 2: Variational Inference}

\begin{algorithm}[!t]
\caption{Variational Inference for CHDP}
\label{ag:vi}
initialization\;
\Do{convergence}{
    Obtain samples of $\prod_a\prod_o q(\nu_{a,o}|u^{(i)}_{a,o}, r^{(i)}_{a,o}) q(z_{a,o}|\varsigma^{(i)}_{a,o,k})$ for \emph{CHDP-Maximization};\\
    \For{$d=1; d \le D$}
    {
        \For{$n=1; 1 \le N_d$}
        {
            Update $\varsigma_{d,n}$ by Eq. (\ref{vi:varsigmadnt});
        }
        Update $u_{d,t}$ and $r_{d,t}$ by Eqs. (\ref{vi:udt}) and (\ref{vi:rdt}); \\
        //\emph{CHDP-Superposition} \\
        Update $\varsigma_{d,t}$ by Eq. (\ref{vis:varsigmadtao});\\
        //\emph{CHDP-Maximization} \\
        Update $\varsigma_{d,t}$ by Eq. (\ref{vim:varsigmadtao});
    }

    \For{$a=1; a \le A$}
    {
        //\emph{CHDP-Superposition} \\
        Update $u_{a,o}$ and $r_{a,o}$ using derivatives in (\ref{vis:uao}) and (\ref{vis:rao});  \\
        Update $\varsigma_{a,o,k}$ using derivative in (\ref{vis:varsigmaaok});  \\
        //\emph{CHDP-Maximization} \\
        Update $u_{a,o}$ and $r_{a,o}$ using derivatives in (\ref{vim:uao}) and (\ref{vim:rao}); \\
        Update $\varsigma_{a,o,k}$ using derivative in (\ref{vim:varsigmaaok});
    }

    \For{$k=1; k \le K^{\dag}$}
    {
        Update $u_{0,k}$ and $r_{0,k}$ by Eqs. (\ref{vi:u0k}) and (\ref{vi:r0k});\\
        Update $\vartheta_{k,v}$ by Eq. (\ref{vi:varthetakv});
    }
}
return $K$, $\{\vartheta_k\}$, $\{u_0, r_0\}$, $\{u_a, r_a\}$, $\{u_d, r_d\}$, $\{\varsigma_{a,o}\}$, $\{\varsigma_{d,t}\}$, $\{\varsigma_{d,n}\}$\;
\end{algorithm}

\textbf{Update $\vartheta_{k,v}$.} The derivative of $\pounds(q)$ on $\vartheta_{k,v}$ with additional proximal regularization is
\begin{align*}
\frac{\partial \pounds_{\vartheta}(q)}{\partial \vartheta_{k,v}}
=&\left(\eta_{v} - (1+\gamma)\vartheta_{k,v} + \gamma\vartheta^{(i)}_{k,v} +
\sum_d \sum_n \delta(w_{d,n}=v) \sum_{ao} \varsigma_{ao,k} \sum_t \varsigma_{d,t,ao} \varsigma_{d,n,t}
\right)
\Psi^{\prime}(\vartheta_{k,v})
\\
+  \sum_v &\left(\eta_{v} - (1+\gamma)\vartheta_{k,v} + \gamma\vartheta^{(i)}_{k,v}+ \sum_d \sum_n \delta(w_{d,n}=v) \sum_{ao} \varsigma_{ao,k} \sum_t \varsigma_{d,t,ao} \varsigma_{d,n,t}
\right)
\left(-\Psi^{\prime}(\sum_v \vartheta_{k,v})\right)
\end{align*}
Finally, it can be updated by
\begin{align}
\label{vi:varthetakv}
\vartheta_{k,v} =  \frac{\eta_{v} + \gamma\vartheta^{(i)}_{k,v} + \sum_d \sum_n \delta(w_{d,n}=v) \sum_{ao} \varsigma_{ao,k} \sum_t \varsigma_{d,t,ao} \varsigma_{d,n,t} }{\gamma + 1}
\end{align}
In addition, the inference could be further speed up by using the stochastic gradient method \cite{hoffman2013stochastic}: Each iteration only selects a batch of documents, and update $\vartheta_k$ that is considered as global variables by slightly revising the above equation.

\textbf{Update $u_{0,k}$ and $r_{0,k}$.} The update of variational parameter $u_{0,k}$ and $r_{0,k}$ is by
\begin{align}
\label{vi:u0k}
u^{(i+1)}_{0,k} = \frac{\sum_a \sum_o \varsigma_{a,o,k} + \gamma( u^{(i)}_{0,k}-1)}{1+\gamma}+1
\end{align}
and
\begin{align}
\label{vi:r0k}
r^{(i+1)}_{0,k} = \frac{\alpha_0 -1 + \sum_a \sum_o \sum_{l>k} \varsigma_{a,o,l}  +\gamma(r^{(i)}_{0,k}-1)}{1+\gamma} +1
\end{align}

\textbf{Update $u_{d,t}$ and $r_{d,t}$.} The update of variational parameter $u_{d,t}$ and $r_{d,t}$ are by
\begin{align}
\label{vi:udt}
u^{(i+1)}_{d,t} = \frac{ \sum_n \varsigma_{d,n,t} +\gamma( u^{(i)}_{d,t}-1)}{1+\gamma}+1
\end{align}
and
\begin{align}
\label{vi:rdt}
r^{(i+1)}_{d,t} = \frac{\alpha_d -1 + \sum_n \sum_{l>t} \varsigma_{d,n,l} +\gamma(r^{(i)}_{d,t}-1)}{1+\gamma} +1
\end{align}

\textbf{Update $\varsigma_{d,n,t}$.}  The update of variational parameter $\varsigma_{d,n,t}$ is by
\begin{align}
\label{vi:varsigmadnt}
\varsigma_{d,n,t}^{(i+1)} \propto &\exp \Bigg\{
\frac{1}{1+\gamma}
\Bigg (
\left (\Psi(u_{d,t}) - \Psi(u_{d,t} + r_{d,t}) \right ) + \sum_{j<t}\left(\Psi(r_{d,j}) - \Psi(u_{d,j} + r_{d,j})\right) - (1+\gamma) +\gamma\log\varsigma^{(i)}_{d,n,t}
\\
&~~~~~~~~~~~
+\sum_k \sum_{ao} \varsigma_{a,o,k}
\varsigma_{d,t,ao}  \sum_v \delta(w_{d,n}=v)
\left ( \Psi(\vartheta_{k,v}) - \Psi\left(\sum_v \vartheta_{k,v}\right) \right )
\Bigg )\Bigg\}
\nonumber
\end{align}
When updating $\varsigma_{d,n,T}$, the item, i.e., $\Psi(u_{d,t}) - \Psi(u_{d,t} + r_{d,t})$ should be removed because $\nu_{d, T} =1$.

\end{document}